\documentclass{article} 
\usepackage{iclr2027_conference,times}
\iclrfinalcopy

\usepackage{amsmath,amsfonts,bm}

\def\eqref#1{equation~\ref{#1}}

\def\1{\bm{1}}

\DeclareMathAlphabet{\mathsfit}{\encodingdefault}{\sfdefault}{m}{sl}
\SetMathAlphabet{\mathsfit}{bold}{\encodingdefault}{\sfdefault}{bx}{n}

\usepackage{hyperref}
\usepackage{url}

\usepackage{booktabs}   
\usepackage{xcolor}         
\usepackage{natbib}
\usepackage[table]{xcolor}
\usepackage{amsmath}
\usepackage{multirow}
\usepackage{caption}
\usepackage[export]{adjustbox}
\usepackage{tabularx}
\usepackage{wrapfig}
\usepackage{tcolorbox}
\tcbuselibrary{breakable}
\usepackage{inconsolata}
\usepackage{algorithm}
\usepackage{algpseudocode}
\usepackage{makecell}

\tcbset{
    promptbox/.style={
        colback=white,
        colframe=black,
        boxrule=0.5pt,
        fonttitle=\small\bfseries,
        fontupper=\small\ttfamily,
        left=6pt, right=6pt, top=4pt, bottom=4pt,
        before upper={\setlength{\parskip}{4pt}}
    }
}

\title{MMCOMET: A Large-Scale Multimodal Commonsense Knowledge Graph for Visual Reasoning}

\newcounter{eqfn}

\author{
Jinwoo Kim\textsuperscript{1}\thanks{Equal contribution.}\setcounter{eqfn}{\value{footnote}} \And 
Rina Carines Cabral\textsuperscript{1}\footnotemark[\value{eqfn}] \And 
Dhita Putri Pratama\textsuperscript{2}\footnotemark[\value{eqfn}] \And
Eileen Wang\textsuperscript{3}\footnotemark[\value{eqfn}] \And
Hiba Arnaout\textsuperscript{4}\footnotemark[\value{eqfn}] \And 
Shuo Yang\textsuperscript{1} \And  Danyang Liu\textsuperscript{5} \And Seungyeon Ji\textsuperscript{6} \And Jie Yang\textsuperscript{7} \And Josiah Poon\textsuperscript{7} \And Jeff Z. Pan\textsuperscript{8} \And Kyungreem Han\textsuperscript{6,9} \And Soyeon Caren Han \textsuperscript{1}\thanks{Corresponding authors: Soyeon Caren Han~\texttt{<caren.han@unimelb.edu.au>}} \AND
\mdseries
\begin{tabular}[t]{@{}c@{}}
\textsuperscript{1}The University of Melbourne \quad
\textsuperscript{2}Mekari \quad
\textsuperscript{3}Canva \quad
\textsuperscript{4}Technical University of Darmstadt \\
\textsuperscript{5}Reka AI \quad
\textsuperscript{6}Korea Institute of Science and Technology \quad
\textsuperscript{7}The University of Sydney \\
\textsuperscript{8}The University of Edinburgh \quad
\textsuperscript{9}Division of Biomedical Science \& Technology \\
\end{tabular}
}

\begin{document}

\maketitle

\begin{abstract}
\vspace{-1em}
Commonsense knowledge is widely used to support multimodal reasoning and generation, yet existing resources are predominantly text-only and lack grounding in visual context. This limitation often leads to the retrieval of semantically related but visually misaligned knowledge, and can degrade downstream performance. 
We introduce \textbf{MMCOMET}, a large-scale multimodal commonsense knowledge graph that integrates physical, social, and eventive knowledge with visual grounding. MMCOMET extends ATOMIC2020 by associating each commonsense triple with representative images through a hybrid alignment pipeline, resulting in over 900K multimodal tuples spanning 19 relation types.
We further establish a unified evaluation framework across diverse vision-language tasks, including visual commonsense reasoning, visual question answering, visual storytelling, image editing, and image captioning. Our analysis shows that multimodal grounding provides more contextually aligned and reliable knowledge compared to both text-only knowledge graphs and LLM-based knowledge generation, particularly in reasoning-intensive scenarios. At the same time, we observe that the effectiveness of external knowledge is task-dependent, and naive knowledge integration can introduce noise in perception-dominated settings. To mitigate this, we propose an adaptive knowledge router that dynamically selects optimal paths or bypass knowledge in harmful settings.
MMCOMET provides a scalable multimodal commonsense resource and a benchmark for studying the role of grounded knowledge in multimodal systems.\footnote{Code and data are available at \url{https://anonymous.4open.science/r/MMCOMET_CODE-4595}}
\end{abstract}

\vspace{-1.5em}
\section{Introduction}
\label{sec:intro}
\vspace{-0.8em}

Commonsense knowledge graphs (KGs) encode everyday knowledge as relational triples
$\langle \texttt{head}, \texttt{relation}, \texttt{tail} \rangle$
and have become important external resources for retrieval-augmented reasoning and generation~\citep{guu2020retrieval}.
In multimodal settings, such knowledge provides auxiliary context for tasks including visual commonsense reasoning (VCR)~\citep{mmcomet_vcr}, visual question answering (VQA)~\citep{garderes2020conceptbert}, visual storytelling (VST)~\citep{wang2024sco}, image editing (IE)~\citep{mmcomet_piebench}, and image captioning (IC)~\citep{zhou2019improving}.
In these pipelines, the quality and relevance of retrieved knowledge directly affect downstream outputs~\citep{kgquality}.
However, most existing commonsense KGs remain text-only and therefore cannot directly support visually grounded knowledge retrieval.

\begin{figure*}
    \centering
    \includegraphics[width=\textwidth]{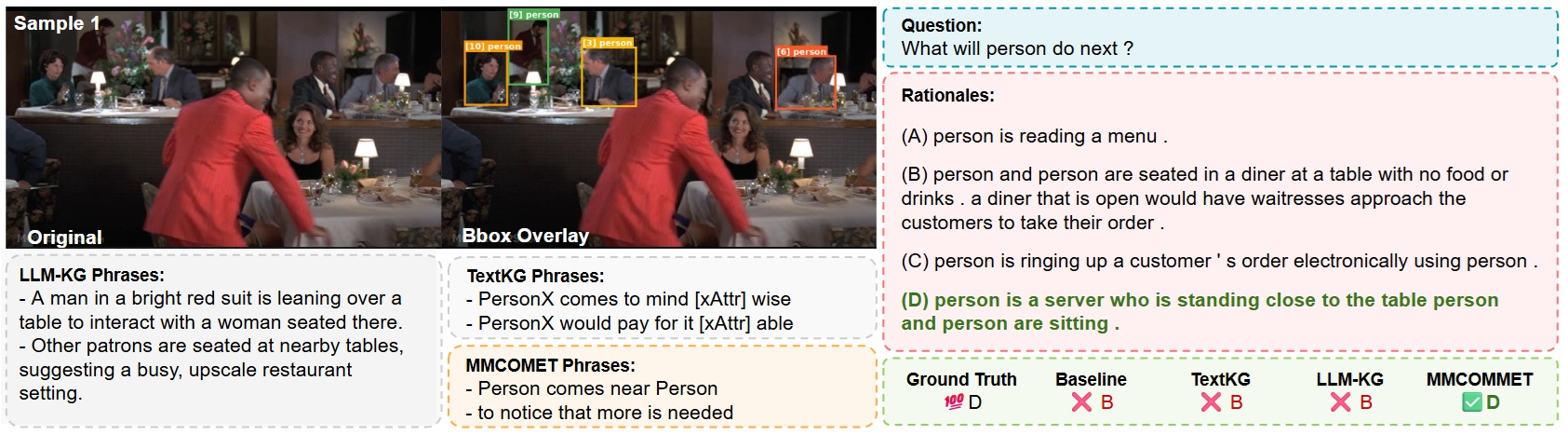}
    \caption{An example of visual commonsense reasoning featuring results from no knowledge injected (baseline), text-based knowledge graph injection (TextKG), LLM-generated knowledge injection (LLM-KG), and MMCOMET knowledge injection.}
    \label{fig:vcr_toy_example}
    \vspace{-2em}
\end{figure*}

Recent multimodal knowledge graphs (MMKGs) incorporate images into large-scale encyclopedic resources~\citep{ferrada2017imgpedia,alberts2020visualsem}.
Although effective for factual and entity-centric queries, they primarily represent encyclopedic relations, such as \textit{located-in} and \textit{capital-of}, and provide limited coverage of everyday commonsense involving social interactions, event dynamics, and human intentions.
Large-scale commonsense KGs such as ATOMIC2020~\citep{hwang2021comet} capture rich social, physical, and eventive knowledge, but remain entirely textual.
Existing resources thus do not jointly represent inferential commonsense and its visual grounding.
Another approach is to generate commonsense knowledge directly using large language models (LLMs).
While flexible, LLM-generated knowledge may be ungrounded, irrelevant, or inconsistent with the visual input.
Similarly, text-only KGs can retrieve knowledge that is semantically related to a query but visually misaligned with the depicted situation.
As illustrated in Figure~\ref{fig:vcr_toy_example}, systems without knowledge or with text-only knowledge may therefore produce generic or weakly grounded reasoning.
Multimodal commonsense can provide more contextually aligned inference by jointly capturing
(i) physical knowledge about objects and environments,
(ii) eventive knowledge about actions and dynamics, and
(iii) social knowledge about roles, intentions, and interactions.

To address this gap, we introduce \textbf{MMCOMET}, a large-scale multimodal commonsense KG that integrates physical, social, and eventive knowledge with visual grounding.
MMCOMET extends ATOMIC2020 by associating commonsense phrases with representative images through a hybrid pipeline that combines embedding-based retrieval with web-scale search.
The resulting resource contains approximately 1M multimodal commonsense tuples spanning 19 relation types.
Beyond constructing the resource, we establish a unified evaluation framework for studying multimodal commonsense across VCR, VQA, VST, IE, and IC.
We compare MMCOMET with text-only KG and LLM-generated knowledge.
Our experiments show that visually grounded commonsense provides more contextually aligned and reliable knowledge, particularly for reasoning-intensive tasks.
At the same time, the benefits of external knowledge vary across tasks, indicating that effective knowledge integration requires adaptive knowledge selection rather than naive integration.

Our contributions are summarized as follows:
\textbf{1)} We introduce MMCOMET, a large-scale multimodal commonsense knowledge graph that integrates physical, social, and eventive knowledge with visual grounding, comprising approximately 1M multimodal tuples.
\textbf{2)} We propose a scalable hybrid alignment pipeline that associates commonsense phrases with representative images using embedding-based retrieval and web search.
\textbf{3)} We provide comprehensive dataset statistics and human evaluation validating the quality and relevance of the resulting image--text alignments.
\textbf{4)} We establish a unified evaluation framework across five vision-language tasks and analyze multimodal commonsense against text-only KG retrieval and LLM-generated knowledge. 
\textbf{5)} We propose an adaptive knowledge router that dynamically selects optimal knowledge paths and prevents harmful interventions when no route provides positive utility.
MMCOMET serves as a reusable resource for multimodal knowledge retrieval, reasoning, and generation, as well as a benchmark for studying grounded commonsense in vision-language systems.

\vspace{-1.2em}
\section{Related Work}
\vspace{-1.2em}

\paragraph{Text-only Commonsense Knowledge Graphs}
Commonsense knowledge graphs (CSKGs) encode everyday knowledge beyond encyclopedic facts. Existing resources have been constructed through crowdsourcing~\citep{liu2004conceptnet,speer2017conceptnet}, web-scale extraction~\citep{tandon2014webchild,nguyen2022refined}, and neural scoring~\citep{zhang2020transomcs}. Among them, ATOMIC and ATOMIC2020~\citep{sap2019atomic,hwang2021comet} provide large-scale inferential knowledge about social interactions, events, and physical concepts, supporting tasks such as dialogue and narrative reasoning~\citep{cai2023improving,wang2024sco}. However, these resources remain text-only and cannot assess whether retrieved commonsense knowledge is visually aligned with a given scene.

\begin{figure*}[t]
  \centering
  \captionof{table}{Comparison of existing MMKGs with MMCOMET (left) and a small subset of MMCOMET (right). MMCOMET uniquely covers domains (social, physical, eventive) with nearly 1M tuples (Ency: Encyclopedic; Geo: Geographic; Multi-L: Multilingual; Edu: Education; CS: Commonsense; So: Social; Ph: Physical; Ev: Eventive; C: Concept; E: Entity; R: Relation; T: Tuples; WSE: Web Search Engine; WP: Wikipedia; WMC: Wikimedia Commons; OID: open-source image datasets; ImgNet: ImageNet.)}
  \vspace{-8pt}
  \begin{minipage}[b]{0.65\textwidth}
    \centering
    \footnotesize
    \begin{tabular}{>{\fontsize{6pt}{7pt}\selectfont}l >{\fontsize{6pt}{7pt}\selectfont}l *{3}{>{\fontsize{6pt}{7pt}\selectfont}l}}
      \hline
        \textbf{MMKG}           & \textbf{Domain} & \textbf{Size} & \textbf{MultiM} & \textbf{Image}\\
        \hline
        IMGpedia~\citep{ferrada2017imgpedia}       & Ency           &     442M & E, R & WMC \\
        ImageGraph~\citep{onoro2017answering}     & Ency        &    560K & E & WSE \\
        MMKG~\citep{liu2019mmkg}       & Ency          &       814K & E &     WSE\\
        MMpedia~\citep{wu2023mmpedia} & Ency & 19M & E & WSE, WP\\
        AspectMMKG~\citep{zhang2023aspectmmkg} & Ency & 645K & E & WSE, WP\\
        UKnow~\citep{uknow_mmkg_2024} & Ency     & 3.6M & E, T & WSE, WP\\
        Richpedia~\citep{wang2020richpedia} & Ency, Geo    &  172M & E, R & WSE, WP   \\
        VisualSem~\citep{alberts2020visualsem}    &     Ency, Multi-L &  1.5M & E & WP, ImgNet       \\
        SciMKG~\citep{scimkg_2026} & Edu & 403K  & C, R, T & WP, OID \\
        MMEKG~\citep{ma2022mmekg} & CS (Ev) & 934M & E, T & imSitu\\
        TIVA-KG~\citep{wang2023tiva} & CS (Ph) & 1.4M & E, T & WSE\\
        VTKG-C~\citep{vista_emnlp_2023} & CS (Ph) & 111K & E, R, T & OID \\
        {\cellcolor{green!15}MMCOMET (\textit{ours})} & {\cellcolor{green!15}CS (So, Ph, Ev)} & {\cellcolor{green!15}989K} & {\cellcolor{green!15}E, R} & {\cellcolor{green!15}WSE, OID}\\
        \hline
    \end{tabular}
  \end{minipage}
  \begin{minipage}[b]{0.33\textwidth}
    \centering
    \includegraphics[width=\textwidth, valign=m]{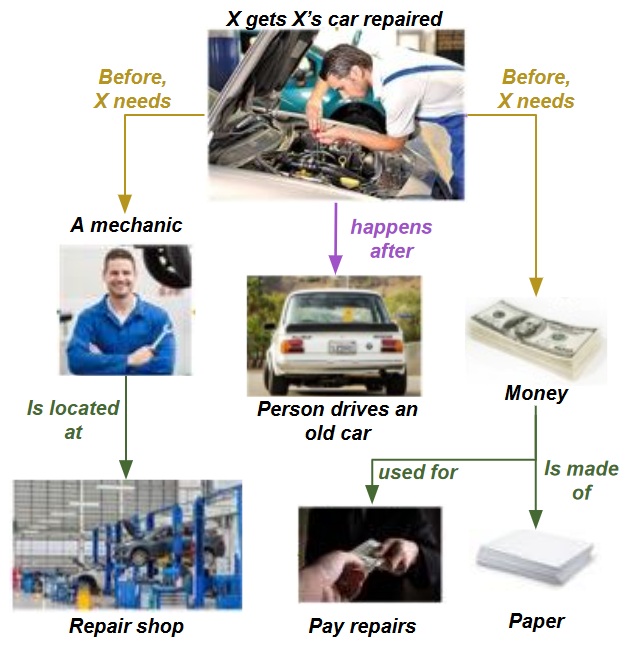}
  \end{minipage}
    \label{tab:comparisontable}
\vspace{-0.8em}
\end{figure*}

\vspace{-1.2em}
\paragraph{Multimodal Knowledge Graphs}
Multimodal knowledge graphs (MMKGs) associate structured knowledge with images or other modalities. Most existing MMKGs ground encyclopedic entities and factual relations from resources such as DBpedia, Freebase, and Wikidata~\citep{ferrada2017imgpedia,liu2019mmkg,wang2020richpedia,wu2023mmpedia,uknow_mmkg_2024}. A smaller set targets commonsense knowledge: MMEKG~\citep{ma2022mmekg} focuses on multimodal events, TIVA-KG~\citep{wang2023tiva} emphasises physical commonsense, and VTKG-C~\citep{vista_emnlp_2023} grounds visually expressible entities and triplets. As summarised in Table~\ref{tab:comparisontable}, these resources do not jointly cover social, physical, and eventive commonsense at scale. MMCOMET addresses this gap by visually grounding nearly 1M ATOMIC2020 triples, providing a unified resource for multimodal commonsense retrieval, reasoning, and generation.

\vspace{-1.2em}
\section{MMCOMET} \label{sec:mmcomet_method}
\vspace{-0.8em}
\subsection{Base Commonsense Topology}
\label{subsec:base_topology}
\begin{figure*}[t]
  \centering
  \includegraphics[width=0.98\textwidth]{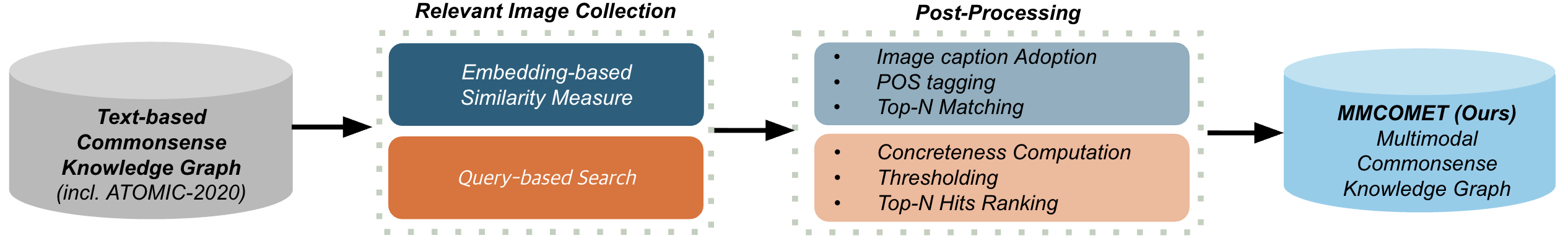}
  \caption{The MMCOMET construction pipeline for grounding a text-based commonsense knowledge graph with representative visual evidence.}
  \label{fig:arch}
  \vspace{-1em}
\end{figure*}

\vspace{-0.8em}
MMCOMET is constructed by extending the textual commonsense graph ATOMIC2020~\citep{hwang2021comet} with a visually grounded layer. For each triple $\langle \texttt{head}, \texttt{relation}, \texttt{tail} \rangle$, we associate representative images for the head and tail phrases through a hybrid retrieval pipeline. Table~\ref{tab:comparisontable} shows a subset of MMCOMET (right). Figure \ref{fig:arch} presents the overall construction workflow.
We adopt ATOMIC2020 as the textual backbone, which contains $\sim$1.33M inferential commonsense triples spanning social, physical, and eventive relations. After filtering incomplete tuples and non-visualizable relation types, including the \texttt{isFilledBy}, which contains blank placeholders, MMCOMET retains 989K multimodal tuples spanning 19 relation categories. MMCOMET does not aim to alter the relational structure, but to \textit{ground} each textual node with representative visual evidence, enabling multimodal reasoning.

\vspace{-0.5em}
\subsection{Hybrid Visual Alignment Pipeline}
\label{subsec:hybrid_alignment}
\vspace{-0.8em}
Associating images with commonsense phrases presents two challenges: (i) physical concepts are often visually concrete and (ii) social and eventive concepts are frequently abstract.
To address this, we design a two-stage retrieval strategy:

\vspace{-0.8em}
\paragraph{Embedding-based Similarity Matching}
For visually concrete phrases (e.g., \texttt{ObjectUse}), we adopt embedding-based retrieval. Head and tail phrases are encoded using CLIP (ViT-B/32)~\citep{radford2021learning} text encoders and matched via cosine similarity against pre-encoded image embeddings from open-source captioning and storytelling datasets Conceptual Captions~\citep{sharma2018conceptual}, COCO~\citep{lin2014microsoft}, Flickr30K~\citep{young2014image}, and the Visual Storytelling dataset~\citep{huang2016visual}.
To ensure scalability, we introduce a noun-indexed filtering mechanism. 
Image captions are POS-tagged using NLTK~\citep{bird2009natural} to extract noun tokens, which are organized into an inverted index, mapping nouns to candidate images. Similarity is first computed at the noun level to identify a reduced candidate subset, after which fine-grained phrase-to-image similarity is applied. 

\vspace{-1.2em}
\paragraph{Concreteness-aware Web Retrieval}
For abstract social and eventive phrases (e.g., \texttt{React}, \texttt{Intent}, \texttt{Want}), caption-based similarity matching often fails to produce adequate visual grounding. 
We employ a concreteness-aware routing strategy.
We compute phrase-level concreteness scores using the 40K Concreteness Ratings~\citep{brysbaert2014concreteness}. Phrases with an average concreteness score below 4.0, a threshold aligned with the empirical distribution of social and eventive relations, are routed to web-based image retrieval. Structured web queries are issued with entity normalization and text-filtering heuristics to reduce noise. Approximately 72\% of unique phrases are handled through this abstract-concept routing process. 

\vspace{-0.5em}
\subsection{Post-processing and Image Selection}
\label{subsec:postprocessing}
\vspace{-0.5em}
All candidate images (from similarity matching and web retrieval) are re-ranked using CLIP similarity with the original phrase. Low-alignment samples are filtered using a threshold of 0.15, and the top 10--15 images per phrase are retained. The final MMCOMET graph consists of 989K multimodal commonsense triples, 578K unique textual nodes, and $\sim$3.9M associated images in social, physical, and eventive relations.
This hybrid pipeline enables scalable multimodal grounding while maintaining high alignment quality, forming a retrieval-ready multimodal commonsense resource. We provide extensive details, statistics, examples, retrieval analysis, and error analysis of the MMCOMET graph construction in the Appendix.~\ref{appendix:multimodal_graph}.

\vspace{-0.5em}

\begin{table}
\centering
\scriptsize
  \caption{
  (Left) Human evaluation results showing the percentage of commonsense phrases whose retrieved images were judged as full (FM) or partial  matches (PM). (Right) Average concreteness scores and sample sizes.
  }
  \vspace{-0.8em}
  \begin{tabular}{@{}l||cc|cc|cc||cc}
    \hline
     & \multicolumn{2}{|c|}{\textbf{Web (WSE)}} & \multicolumn{2}{|c|}{\textbf{Open-source Data (OID)}} & \multicolumn{2}{|c||}{\textbf{Both}} & \multicolumn{2}{c}{\textbf{Relation Type Stats.}} \\\hline \hline
     \textbf{Physical} & \textbf{FM+PM} & \textbf{FM} & \textbf{FM+PM} & \textbf{FM} & \textbf{FM+PM} & \textbf{FM} & \textbf{Avg. Concreteness} & \textbf{Size} \\ \hline
    ObjectUse & 100 & 73 & 96 & 76 & 98 & 74.5 
        & 3.90 & 165,590 \\
    AtLocation & 100 & 92.9 & 100 & 100 & 100 & 96.4 
        & 4.15& 20,221 \\
    MadeUpOf & 100 & 100 & 89.3 & 89.3 & 94.1 & 94.1 
        & 4.08 & 3,345 \\ 
    HasProperty  & 92.6 & 92.6 & 88.5 & 84.6 & 90.6 & 88.7 
        & 3.80 & 5,617 \\
    CapableOf  & 95.8 & 87.5 & 95.8 & 70.8 & 95.8 & 79.2 
        & 3.96 & 7,968 \\
    Desires  & 100 & 95.7 & 100 & 72 & 100 & 83.3 
        & 3.88 & 2,737 \\
    NotDesires  & 100 & 92.6 & 88 & 72 & 94.2 & 82.7 
        & 3.91 & 2,838 \\
    \hline
     \textbf{Event} & \textbf{FM+PM} & \textbf{FM} & \textbf{FM+PM} & \textbf{FM} & \textbf{FM+PM} & \textbf{FM} & \textbf{Avg. Concreteness} & \textbf{Size} \\ \hline
    isAfter & 91.3 & 69.6 & 77.3 & 45.5 & 84.4 & 57.8 
        & 3.94 & 22,453 \\
    HasSubEvent & 96.3 & 92.6 & 92.9 & 71.4 & 94.5 & 81.8 
        & 3.58 & 12,845 \\
    isBefore & 87.5 & 62.5 & 88.9 & 63 & 88.2 & 62.7 
        & 3.91 & 23,208 \\ 
    HinderedBy  & 86.7 & 73.3 & 76.2 & 42.9 & 80.6 & 55.6 
        & 3.80 & 106,647 \\
    Causes  & 96.2 & 96.2 & 87.5 & 79.2 & 92 & 88 
        & 3.29 & 376 \\
    Reason  & 100 & 92 & 83.3 & 75 & 91.8 & 83.7 
        & 3.55 & 334 \\
    \hline
     \textbf{Social} & \textbf{FM+PM} & \textbf{FM} & \textbf{FM+PM} & \textbf{FM} & \textbf{FM+PM} & \textbf{FM} & \textbf{Avg. Concreteness} & \textbf{Size} \\ \hline
    Need & 90.5 & 81 & 85.7 & 71.4 & 88.1 & 76.2 
        & 3.65 & 89,391 \\
    Attr & 92 & 80 & 89.5 & 78.9 & 90.9 & 79.5
        & 2.98   & 115,815 \\
    Effect & 95 & 90 & 61.9 & 42.9 & 78 & 65.9 
        & 3.53 & 113,494 \\ 
    React  & 90.9 & 86.4 & 61.1 & 55.6 & 77.5 & 72.5 
        & 2.95 & 94,228 \\
    Want  & 80.8 & 73.1 & 86.4 & 50 & 83.3 & 62.5 
        & 3.58 & 152,989  \\
    Intent  & 95 & 80 & 88.2 & 58.9 & 91.9 & 70.3 
        & 3.40  & 49,312 \\
    \hline
   \textbf{All Relations} & 91.5 & 79.0 & 82.4 & 61.2  & 86.9 & 70.1 & 3.55 & 989,408\\
    \hline
  \end{tabular}
  \label{tab: human_eval} 
  \vspace{-10pt}
\end{table}
\vspace{-0.5em}
\section{Intrinsic Evaluation of MMCOMET: Multimodal Alignment}
\label{sec:intrinsic_eval}

\vspace{-0.7em}
\paragraph{Human Evaluation}
We conduct a human evaluation to assess the semantic alignment between commonsense phrases and their aligned images in MMCOMET. 
For each of the 19 relations, 60 phrases are randomly sampled, resulting in 1140 phrases divided among three annotators (Appendix~ \ref{appendix:human_eval_info}). To balance retrieval strategies, 50\% of the samples are drawn from phrases that required web search.
For each head/tail phrase, annotators are shown the top 7 images, yielding 7980 phrase-image pairs. Each image is then labeled as either 1) \textit{Full Match (FM)}, where all key concepts expressed in the phrase are visually present; 2) \textit{Partial Match (PM)}, where the image is thematically relevant but missing some elements; or 3) \textit{No Match (NM)}, where the image does not correspond to the phrase. 
Additionally, annotators rate concreteness as \textit{Low}, \textit{Medium}, or \textit{High}, corresponding to the difficulty of visually grounding the phrase. In particular, \textit{Low} denotes abstract or intangible items, while \textit{High} are concrete, tangible objects. Examples for each semantic alignment label may be found in the Appendix. \ref{appendix:full_vs_partial}.

Table~\ref{tab: human_eval} reports the percentage of phrases for which at least 5 out of the top 7 retrieved images were judged as either Full Match (FM) or Partial Match (PM).
Across all relations and image sources, MMCOMET achieves 86.9\% FM+PM alignment and 70.1\% FM alignment, indicating that the majority of commonsense phrases are grounded with visually relevant evidence. Although Full Match scores are naturally lower, partially matched images remain semantically relevant and preserve core contextual elements, making them suitable for downstream retrieval and reasoning.
Inter-annotator agreement is strong, with intraclass correlation coefficients of 0.76 (FM+PM) and 0.82 (FM), indicating consistent annotation quality.
Overall, the human evaluation confirms that MMCOMET provides high-quality multimodal grounding across diverse commonsense relations, supporting its reliability as a retrieval-oriented multimodal resource.

\noindent \textbf{Concreteness Analysis by Relation Type}
We analyze phrase-level concreteness scores across relation types to better understand grounding difficulty and the necessity of hybrid retrieval. The right sub-table of Table~\ref{tab: human_eval} reports average concreteness scores per relation.
Physical relations exhibit consistently higher concreteness scores than social and eventive relations, with the exception of temporal relations such as \texttt{isBefore} and \texttt{isAfter}. Social-interaction relations show the lowest average concreteness, reflecting their reliance on abstract emotional and behavioral descriptions. For example, \texttt{React} frequently involves affective expressions, while \texttt{Attr} often consists of adjectival characterizations, both of which are inherently difficult to ground visually.
We provide further analysis and examples in the Appendix. \ref{appendix:qa_intrinsic}.

\section{Extrinsic Validation of MMCOMET: Downstream Task Evaluation}
\label{sec:experimental_results}
\vspace{-0.5em}
To assess the practical utility of MMCOMET as a reusable multimodal commonsense resource, we evaluate its impact on five vision-language tasks: Visual Commonsense Reasoning (VCR), Visual Question Answering (VQA), Visual Storytelling (VST), Image Editing (IE), and Image Captioning (IC)\footnote{Across these tasks, we show that injecting MMCOMET's multimodal commonsense knowledge improves performance. However, this benefit is not uniform across all task types (See Sec.~\ref{sec:router}). We therefore do not claim that MMCOMET is broadly helpful for any vision-language task, but rather that it is most effective when a task requires commonsense-driven reasoning beyond direct visual perception.}. For each task, we provide extensive examples and qualitative analysis in the Appendix.~\ref{appendix:qa_extrinsic}.

\vspace{-0.8em}
\paragraph{Experimental Setup}
For all tasks except Visual Storytelling, we compare three knowledge injection setups: 1) \textbf{Baseline} with no knowledge injected; 2) \textbf{TextKG}, which injects ATOMIC2020 text-based knowledge triplets into each task prompt; and 3) \textbf{MMCOMET}, which injects multimodal commonsense knowledge phrases incorporating physical, social, and eventive dimensions. We inject the top $k$ knowledge triplets/phrases retrieved using L2 similarity between the image, the triplet/phrase, and the query for each task sample. We empirically set best $k$, where the performance is consistently strong across tasks. As the retrieval unit is identical across conditions, TextKG and MMCOMET receive an equal amount of injected knowledge. Exploring task-specific values of $k$ remains a direction for future work.
We evaluate Visual Storytelling by enriching the story graph with MMCOMET and compare it with prior work (Section \ref{sec:extrinsic_vst}). All experiments were conducted with single H100 GPU with 94GB GPU memory.
Task-specific datasets and metrics are further discussed in each task section. Relevant model details, hyperparameters, and dataset statistics may be found in Appendix~\ref{appendix:eval_setup}. Prompt templates are provided in Appendix~\ref{appendix:prompts}.

\vspace{-0.5em}

\subsection{Visual Commonsense Reasoning (VCR)}
\vspace{-0.8em}
\begin{wrapfigure}{R}{0.5\textwidth}
    \centering
    \scriptsize  
    \captionof{table}{VCR (Q$\!\rightarrow\!$ R) accuracy with different settings. $\Delta$ values represents performance differences between TextKG (T) and MMCOMET (M) compared to the Baseline (B). MMCOMET-enhanced models outperform baselines in accuracy in both smaller and larger models.}
    \begin{tabular}{lccccc}
      \toprule
      \textbf{Model}
        & \textbf{B} & \textbf{T}
        & \textbf{M} & $\boldsymbol{\Delta}_{T-B}$ & $\boldsymbol{\Delta}_{M-B}$ \\
        \midrule
      Qwen8B
        & 65.3 & 62.7 & \textbf{66.4} & -2.6 & +1.1 \\
      Qwen32B
        & 72.1 & 71.1 & \textbf{72.4} & -1.0 & +0.3 \\
      InternVL38B
        & 73.8 & 74.1 & \textbf{75.3} & +0.3 & +1.5 \\
      \bottomrule
    \end{tabular}
    \label{tab:results_vcr}
    \vspace{-15pt}
\end{wrapfigure}
Visual Commonsense Reasoning is a vision-language task that requires models to select a correct answer given a question (Q$\rightarrow$ A) or to select a rationale given a question-answer pair (QA$\rightarrow$ R). In this work, we intentionally modify the task setup to require models to select a rationale from the question alone (Q$\rightarrow$R), without access to the ground-truth answer, encouraging more implicit reasoning and preventing shortcut rationale selection.
We evaluate the significance of multimodal physical, social, and eventive commonsense knowledge to baselines using the VCR benchmark dataset~\citep{mmcomet_vcr}, measure performance using Accuracy, and run experiments with Qwen3-VL variants and InternVL3 in zero-shot and few-shot settings. We report the best setups in Table~\ref{tab:results_vcr}.


\paragraph{Overall Results} Across all model setups, TextKG injection degrades performance in Qwen3 models and yields marginal gains in InternVL3 (-2.7 to +0.3), suggesting that TextKG-retrieved knowledge, without visual grounding, tends to introduce noise that misleads some models. MMCOMET shows greater improvements over the baseline across all models (+0.3 to +1.5), suggesting that multimodal grounding, with physical, social, and eventive knowledge, reduces noise and leads to more stable, more reliable reasoning performance than text-only commonsense knowledge.
We further observe that performance gains vary between model architectures and size. For Qwen3, despite having a lower baseline performance, the 8B variant shows greater gains than its 32B counterpart. On the other hand, InternVL3-38B shows the best baseline performance along with the highest gains. These suggest that MMCOMET is more robust across different model architectures and sizes, while also indicating the need to calibrate the injected knowledge to the specific model setup.

\vspace{-0.8em}
\paragraph{Ambiguity Analysis} To further investigate when multimodal commonsense knowledge is most beneficial, we analyze model performance based on ambiguity, defined as $Ambiguity = 1\space-(p(top_1)\space - p(top_2))$, where $p(top_1)$ and $p(top_2)$ denote the probabilities assigned to the top-1 and -2 answer candidates by the baseline model, respectively. A higher ambiguity score indicates that the model assigns similar probabilities to multiple candidates, reflecting samples where the correct answer is less distinguishable, or inherently harder samples. We find that MMCOMET improves accuracy and reduces uncertainty, outperforming Baseline and TextKG across different categories described in Figure~\ref{fig: ambiguity_accuracy}. Further, the performance gap between MMCOMET and TextKG grows as ambiguity increases. TextKG becomes detrimental as sample difficulty increases, suggesting that visual grounding is critical for harder samples. More analysis may be found in Appendix~\ref{appendix:ablation_vcr}.
\vspace{-0.8em}
\paragraph{Qualitative Analysis}
Figure~\ref{fig:vcr_toy_example} shows a representative VCR example comparing different external knowledge injected into a  Q$\rightarrow$ R prompt. Across all other setups, only MMCOMET commonsense knowledge achieves the correct rationale for the question "What will person do next?" The TextKG-retrieved knowledge fails to associate any commonsense knowledge relevant to the events depicted in the image, whereas LLM-generated knowledge focuses more on general descriptions of the scene. MMCOMET's social and event-based commonsense knowledge enables it to generate more event-based knowledge ("Person comes near Person" and "to notice that more is needed"), which leads the model to the correct rationale. More examples may be found in Appendix~\ref{appendix:qa_vcr}.

\vspace{-0.8em}
\paragraph{Category Breakdown}
\begin{figure}[!t]
\vspace{-1em}
    \centering    \includegraphics[width=\textwidth]{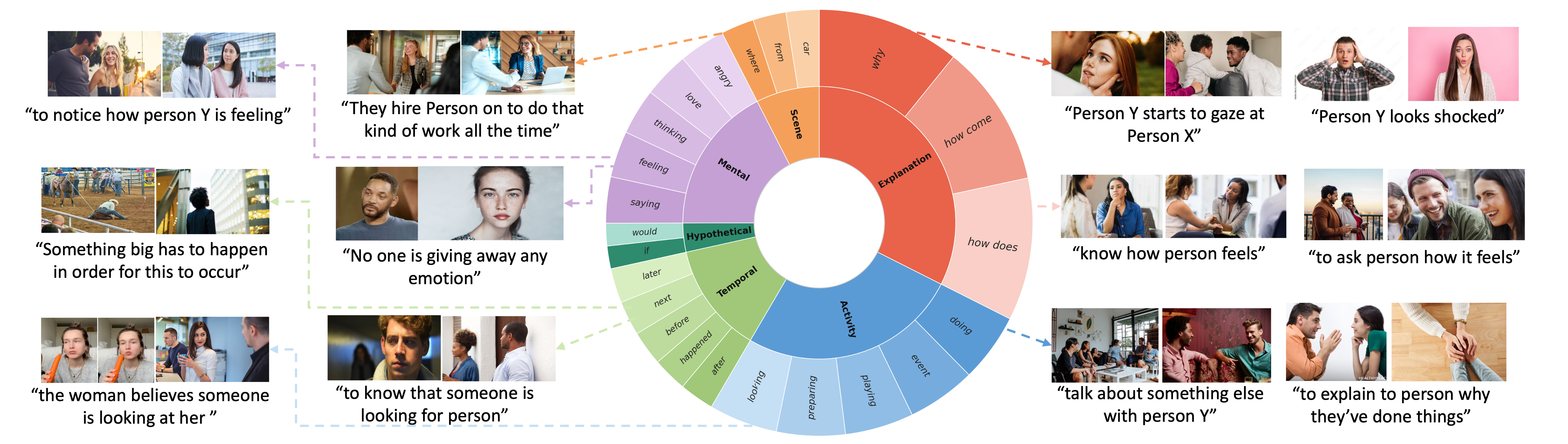}
    \caption{Category breakdown from the VCR dataset with representative commonsense phrases and images from MMCOMET, highlighting the diversity of situational applications of our multimodal physical, social, and eventive commonsense knowledge.}
    \label{fig:vcr_categories}
    \vspace{-15pt}
\end{figure}
We further demonstrate the extrinsic diversity of commonsense knowledge in MMCOMET by retrieving the top MMCOMET images and phrases for different categories in the VCR dataset. Figure~\ref{fig:vcr_categories} illustrates that MMCOMET can visually ground representative images for phrases relevant to each category, such as "No one is giving away any emotion" under \textit{Mental-feeling} and "talk about something else with person Y" under \textit{Activity-doing}. This highlights the versatility of MMCOMET across different situational, social, and eventive applications and across both visual and textual modalities. Further breakdown analysis is in Appendix~\ref{appendix:ablation_vcr}.

\vspace{-1.0em}
\subsection{Visual Question Answering (VQA)}
\vspace{-0.6em}


\begin{wrapfigure}[18]{r}{0.5\textwidth}
    \vspace{-12pt}
    \centering
    \scriptsize
    \captionof{table}{Automated evaluation results and a qualitative sample for VQA. MMCOMET-enhanced models outperform baselines on accuracy across different models.}
    \begin{tabularx}{\linewidth}{l *{3}{>{\centering\arraybackslash}X}}
        \toprule
        \textbf{Model} & \textbf{Baseline} & \textbf{TextKG} & \textbf{MMCOMET} \\
        \midrule
        Qwen8B & 68.60 & 68.41 & \textbf{69.74} \\
        gemma31B & 64.72 & 65.55 & \textbf{66.31} \\
        \bottomrule
    \end{tabularx}    
    \includegraphics[width=0.5\textwidth]{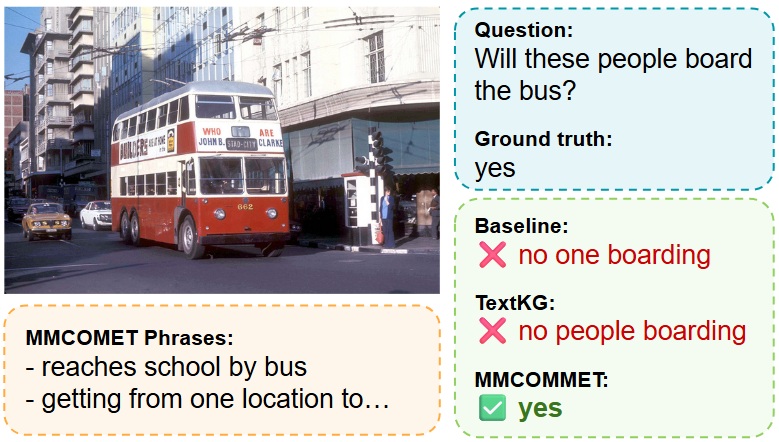}
    \vspace{-10pt}
    \label{tab:results_vqa}

\end{wrapfigure}

Visual Question Answering is a multimodal task in which the primary goal is to answer a question based on a given image. It requires an understanding of visual components and grounding to the textual semantics of the question. External knowledge tends to improve this alignment. We evaluate the significance of our multimodal commonsense knowledge to the VQA task using MLLM models Qwen3 and Gemma, the VQAv2~\citep{vqa_v2_2017} validation dataset, and VQA accuracy~\citep{manas2024improving}, which is a metric robust to inter-human variability. Compared to VCR, VQAv2 consists of more open-ended, perceptual questions that require physical or eventive commonsense knowledge rather than purely complex social reasoning.

Table~\ref{tab:results_vqa} illustrates that injecting MMCOMET consistently improves performance over the baseline across both models, while TextKG yields either marginal gains or slight degradation. This gap suggests that visual grounding is critical even for open-ended factual questions, where TextKG may introduce contextually irrelevant information. The qualitative sample further illustrates the benefit of injecting the eventive knowledge ``getting from one location to another" in answering the given questions. These results indicate that MMCOMET generalises beyond social and intentional reasoning, providing reliable commonsense augmentation across a broader range of question types.



\vspace{-1em}
\subsection{Visual Storytelling (VST)}
\label{sec:extrinsic_vst}

\begin{wrapfigure}[20]{r}{0.5\textwidth}
\vspace{-13pt}

    \centering    
    \captionof{table}{Automatic evaluation results of visual storytelling. MMCOMET-enhanced models outperform baselines in visual grounding, coherence, and overall story quality metrics}
    \scriptsize
        \begin{tabular}{lccc}
    \toprule
     \textbf{Model} & \textbf{RoViST(R)} & \textbf{RSBMU} & \textbf{StoryLength} \\ \midrule
    SCO-VIST & 77.7  & 263.5 & 52.3 \\ 
    VILT
    & 80.2 & 259.2 & 50.1 \\
    LLaVa
    & 82.2 & 263.9 & 51.9 \\
    \-    +MMCOMET$^{S}$ & \underline{82.5} & \underline{264.4} & 50.8 \\ 
    \-    +MMCOMET$^{S+I}$ & \textbf{83.5}  & \textbf{272.0} & 53.4 \\
    \bottomrule
    \end{tabular}
    \qquad
    \includegraphics[width=0.5\textwidth]{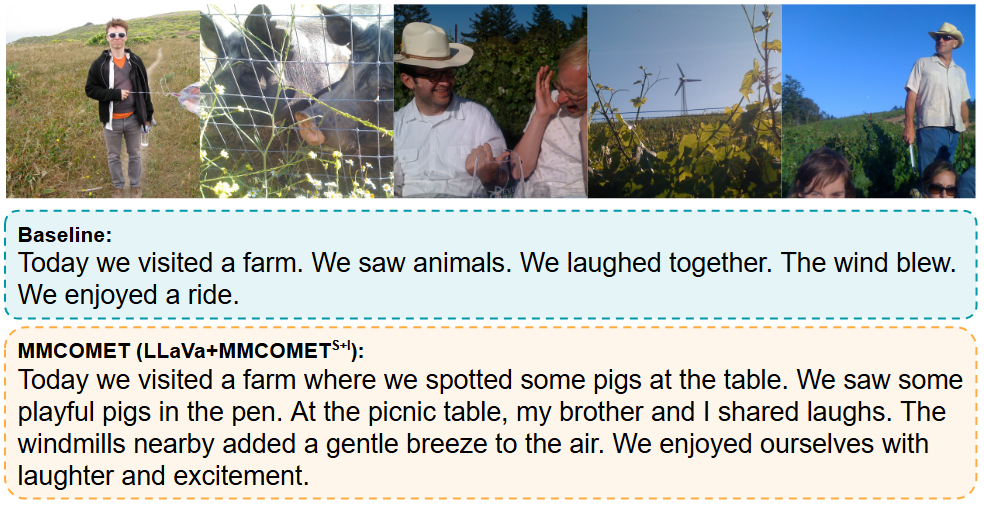}
    
    \label{tab:results_vst_combined}
    \vspace{-15pt}
\end{wrapfigure}

\vspace{-0.5em
}
Unlike the previous tasks, Visual Storytelling~\citep{huang2016visual} requires synthesising a sequence of images to generate a coherent, cohesive storyline, often going beyond factual descriptions and requiring 
interpretation of visual social cues. 
We evaluate MMCOMET using the VIST benchmark~\citep{huang2016visual}, following SCO-VIST's framework~\citep{wang2024sco}.
SCO-VIST's textual story graph is encoded into a multimodal graph by querying MMCOMET to find a matching image for each textual node. 
The story is then decoded by feeding the storyline, MMCOMET and VIST images into LLaVA~\citep{liu2024visual}. 
We compare our setup with the best model from SCO-VIST and a modified setup using VILT~\citep{kim2021vilt} as decoder. 
We measure performance using Perplexity, UNION (U)~\citep{guan2020union}, RoViST~\citep{wang2022rovist} set,
SPICE (S)~\citep{anderson2016spice}, BLEURT (B)~\citep{sellam2020bleurt}, and MoverScore (M)~\citep{zhao2019moverscore}. Averaged RoVist (R) and composite R+S+B+M+U are reported as integrated measures. More details 
may be found in Appendix~\ref{mmcomet-vst}.

Table~\ref{tab:results_vst_combined} 
illustrates that incorporating multimodal commonsense knowledge improves visual grounding, coherence, non-redundancy and semantic similarity (RoViST). LLaVA+MMCOMET$^{S+I}$ achieves the highest overall composite score (272.0) and RoViST (83.5\%), outperforming non-augmented baselines. 
Qualitative examples show that MMCOMET-enhanced stories exhibit richer event descriptions, clearer social interactions, and more contextual details compared to baseline. More samples and results may be found in Appendix~\ref{appendix:qa_vst} and ~\ref{appendix:add_results_vst}.

\vspace{-1.2em}
\subsection{Image Editing (IE)}


\vspace{-0.5em}
\begin{wrapfigure}[19]{r}{0.5\textwidth}
    \vspace{-12pt}
    \centering
    \captionof{table}{Automatic evaluation results and a qualitative sample for image editing. MMCOMET outperforms baselines in SSIM and LPIPS, achieves competitive PSNR, and demonstrates more faithful edits to the source and the instruction.
    }
    \vspace{-1em}
    \scriptsize
    \begin{tabularx}{\linewidth}{l *{3}{>{\centering\arraybackslash}X}}
        \hline
            \toprule
            \textbf{Metric} & \textbf{Baseline} & \textbf{TextKG} & \textbf{MMCOMET} \\
            \midrule
            PSNR $\uparrow$ & 25.7 & \textbf{29.0025} & 28.9692 \\
            SSIM $\uparrow$& 0.86 & 0.9045 & \textbf{0.9273}\\
            LPIPS $\downarrow$ & 0.088 & 0.0728 & \textbf{0.0594} \\
            \bottomrule
        \hline
    \end{tabularx}    \includegraphics[width=0.5\textwidth]{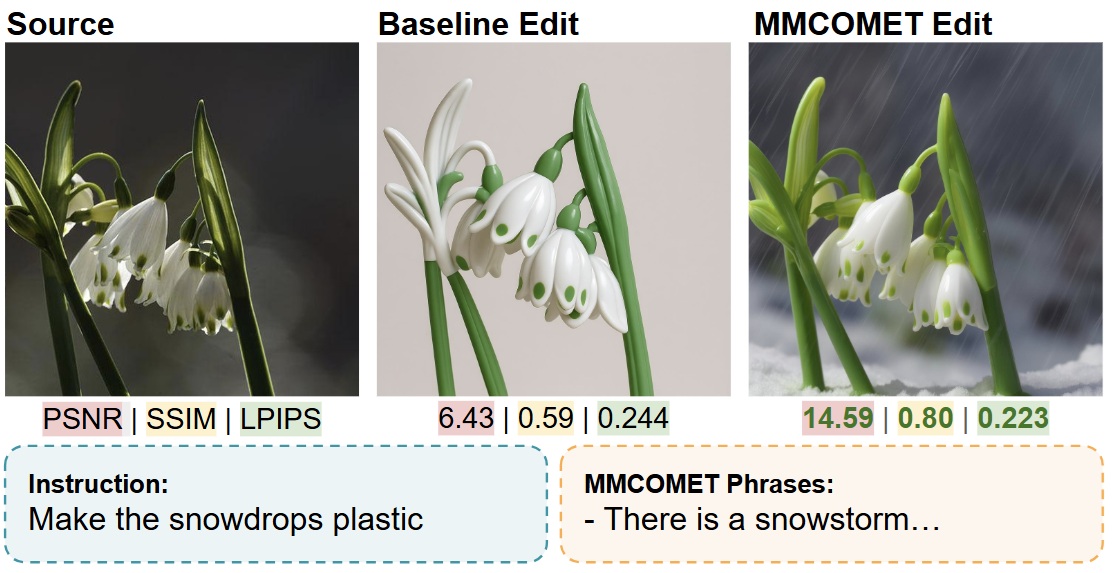}
    \label{tab:result_ie}
\end{wrapfigure}
Image editing is the manipulation of an input image based on a given instruction. The task requires understanding the intent of the edit instruction and visually ground it to the input before generating a new image. To evaluate whether multimodal commonsense improves semantic consistency across various editing categories, we use the PIE\_Bench dataset~\citep{mmcomet_piebench} and Qwen-Image-Edit~\citep{wu2025qwen}.
We measure edited image quality using PSNR, SSIM, and LPIPS.
Table~\ref{tab:result_ie} illustrates that commonsense knowledge improves editing quality over Baseline. MMCOMET further improves gains from TextKG for SSIM (+0.0228) and LPIPS (-0.0134). 
We also provide qualitative examples (Appendix \ref{appendix:qa_ie}) to assess the semantic consistency beyond image quality. 
While the baseline captures the intent of the instruction "Make the snowdrops plastic", it has changed the context of the source image and even removed the background. On the other hand, the MMCOMET edit retains the semantic attributes of the environment shown in the source image through the retrieved knowledge "There is a snowstorm...". Overall, MMCOMET not only improves the quality of edited images but also further establishes semantic consistency between the source and manipulated images.

\vspace{-1.0em}
\subsection{Image Captioning (IC)}

\begin{wrapfigure}[21]{R}{0.48\textwidth}
    \vspace{-10pt}

    \centering    
    \captionof{table}{Automatic evaluation results and a qualitative sample for Image Captioning. MMCOMET-enhanced models outperform other settings in all metrics.}
    \vspace{-5pt}
    \scriptsize
        \begin{tabular*}{\linewidth}{@{\extracolsep{\fill}}lcccc}
        \toprule
      \textbf{Model} & \textbf{Metric}
        & \textbf{B} & \textbf{T} & \textbf{M} \\
      \midrule
      \multirow{3}{*}{\shortstack[c]{Qwen8B}}
        & BLEU-4       & 0.147 & 0.169 & \textbf{0.170} \\
        & CIDEr      & 0.232 & 0.346 & \textbf{0.372} \\
        & ROUGE-L & 0.422 & 0.442 & \textbf{0.446} \\
      \midrule
      \multirow{3}{*}{\shortstack[c]{Qwen32B}}
        & BLEU-4       & 0.139 & 0.143 & \textbf{0.146} \\
        & CIDEr     & 0.151 & 0.147 & \textbf{0.157} \\
        & ROUGE-L & 0.409 & 0.411 & \textbf{0.413} \\
      \midrule
      \multirow{3}{*}{\shortstack[c]{gemma31B}}
        & BLEU-4       & 0.255 & 0.257 & \textbf{0.260} \\
        & CIDEr     & 0.646 & 0.647 & \textbf{0.650} \\
        & ROUGE-L & 0.526 & 0.527 & \textbf{0.530} \\
      \bottomrule
    \end{tabular*}
    \qquad
    \includegraphics[width=0.48\textwidth]{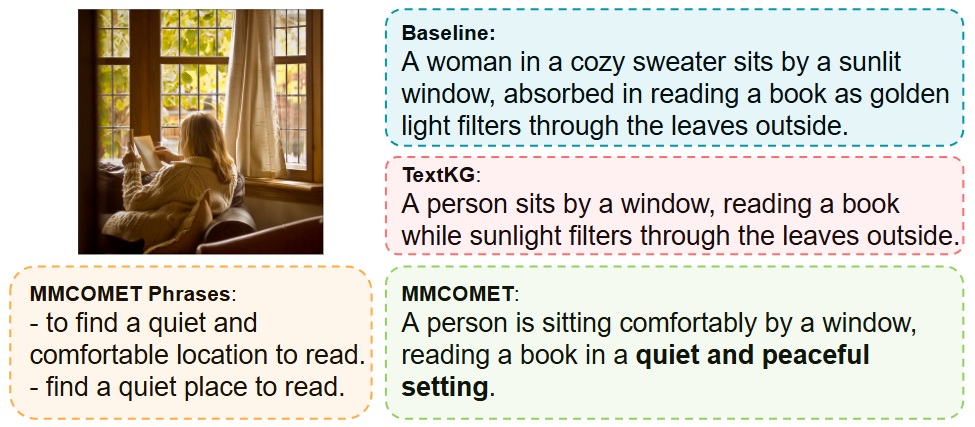}
    \label{tab:results_ic}
    \vspace{-30pt}
\end{wrapfigure}

\vspace{-0.8em}
Image captioning involves generating textual descriptions based on an input image. It requires identifying primary subject/s and understanding context of the image~\citep{hossain2019comprehensive}, which could benefit from multimodal commonsense.
We assess task performance using the Karpathy test split~\citep{karpathy2015deep} of Flickr30k~\citep {young2014image}, a widely used image captioning benchmark, across BLEU, CIDEr, and ROUGE-L metrics with fewshot prompting.

Table~\ref{tab:results_ic} illustrates that knowledge-augmented setups consistently outperform baselines across all metrics. Further, incorporating MMCOMET's commonsense knowledge outperforms TextKG knowledge, highlighting the benefits of multimodal commonsense. Improvements in CIDEr indicate better alignment with the reference descriptions, while ROUGE-L gains suggest semantic richness. The qualitative sample demonstrates that MMCOMET introduces socially and eventively grounded interpretations (e.g. the intent of ``finding a quiet place to read"), leading to more informative and context-aware captions.



\vspace{-0.8em}
\subsection{Extrinsic Human Evaluation} 
\vspace{-0.8em}
Beyond automatic metrics, we conduct a human evaluation across four tasks (VCR, VQA, Image Editing, Image Captioning) to assess whether the retrieved commonsense knowledge is perceived as relevant and useful. Annotators rated 20 samples per task on a 4-point Likert scale for \textbf{(1) relevance of the retrieved knowledge to the image and input}, and \textbf{(2) its helpfulness for reasoning.} Average scores were 3.1 and 2.9, respectively, both above the neutral midpoint, and annotators perceived the retrieved knowledge as helpful even in cases where MMCOMET did not outperform the baseline on automatic metrics, suggesting a gap between automatic evaluation and human judgment. Full annotation details are provided in the Appendix~\ref{appendix:human_eval_info_extrinsic}.

\vspace{-1em}
\subsection{Effect of Knowledge Source}
\label{sec:llm_ablation}
\vspace{-0.8em}
To further demonstrate the effectiveness of MMCOMET, we compare it against two alternative knowledge sources: LLM-generated commonsense and VTKG-C~\citep{vista_emnlp_2023}, an existing multimodal physical knowledge graph.

For LLM-generated commonsnese, we prompt InternVL8B to generate commonsense knowledge phrases about objects, people, relationships, behaviors, and the contextual background of an input image, using 100 samples per task, and inject this in the same setup as MMCOMET. Figure~\ref{fig:ablation_llm_gen} (Appendix~\ref{appendix:llm_gen_figure}) highlights the effectiveness of MMCOMET's multimodal commonsense knowledge for reasoning-dominated tasks. In particular, MMCOMET outperforms all other setups for VQA and VCR, where model reasoning benefits from the integrated physical, social, and eventive knowledge. Notably, LLM-generated knowledge performs worst across all tasks, indicating that the structured knowledge from the commonsense graphs better directs visual grounding than naive knowledge integration. Perception-dominated tasks such as IC and IE benefit less from commonsense knowledge integration; however, greater performance gains are still achieved over text-only knowledge, demonstrating more robust knowledge. More detailed results and comparison to VTKG-C are provided in the Appendix~\ref{appendix:llm_gen_figure}

\vspace{-1.2em}
\section{Knowledge Utility Estimation and Selection with MMCOMET}
\label{sec:router}
\vspace{-0.8em}
\begin{wrapfigure}[16]{r}{0.5\textwidth}
    \vspace{-13pt}
    \centering
    \scriptsize
    \captionof{table}{Adaptive Router (R) on MMStar against baseline and full MMCOMET (M). MMCOMET with a router generally improves upon the baseline or mitigates degradation.}
    \label{tab:router_main}
    \vspace{-10pt}
    \begin{tabular}{@{}llcccc@{}}
    \toprule
    \textbf{Model} & \textbf{Setup} 
    & {\textbf{Overall}} 
    & {\textbf{LR}} 
    & {\textbf{CP}} 
    & {\textbf{MA}} \\ \midrule
    Qwen32B & Baseline & 0.7160 & 0.6912 & 0.7614 & 0.7500 \\
     & M (Full) & 0.7000 & 0.7206 & 0.7500 & 0.7024 \\
     & M + R & \textbf{0.7260} & \textbf{0.7353} & \textbf{0.7841} & \textbf{0.7619} \\ \midrule
    
    LLaVa34B & Baseline & 0.5060 & 0.4265 & \textbf{0.6364} & \textbf{0.4881} \\
     & M (Full) & 0.4720 & 0.4706 & 0.6023 & 0.4405 \\
     & M + R & \textbf{0.5080} & \textbf{0.5000} & \textbf{0.6364} & 0.4762 \\ \midrule
    
    Mistral24B & Baseline & \textbf{0.6020} & 0.5735 & 0.6818 & 0.6071 \\
     & M (Full) & 0.5840 & 0.5441 & 0.6932 & 0.5952 \\
     & M + R & \textbf{0.6020} & \textbf{0.5882} & \textbf{0.7045} & \textbf{0.6310} \\ \midrule
    
    Qwen8B & Baseline & 0.6460 & 0.6176 & \textbf{0.7386} & 0.6548 \\
     & M (Full) & 0.6300 & 0.6029 & 0.7045 & 0.6548 \\
     & M + R & \textbf{0.6520} & \textbf{0.6324} & 0.7159 & \textbf{0.6905} \\ 
    
     \bottomrule
    \end{tabular}
\end{wrapfigure}

Naively integrating all available external knowledge can be harmful when the retrieved knowledge is irrelevant to an instance. We therefore propose an adaptive knowledge router that selects either one of the available MMCOMET knowledge routes or a no-knowledge route. The router predicts an instance-level utility score based on the estimated change in prediction margin relative to the baseline.
It is trained to preserve routes associated with positive gains while bypassing knowledge when predicted utility is insufficient. 
Thus, the router treats knowledge integration as a selective decision rather than a fixed augmentation.

We evaluate the router on MMStar~\citep{chen2024are}, a comprehensive VQA benchmark with six heterogeneous question categories. The candidate set contains seven MMCOMET routes, comprising individual commonsense domains and their combinations. This setting tests whether selective integration retains useful knowledge while avoiding the degradation caused by indiscriminate augmentation.
Table~\ref{tab:router_main} shows that applying all MMCOMET knowledge uniformly (\textit{M (Full)}) substantially reduces performance relative to the baseline. In contrast, MMCOMET with the adaptive router (\textit{M + R}) consistently improves overall performance over both the baseline and full knowledge integration. The largest gains occur on Logical Reasoning (\textit{LR}), particularly for larger models, indicating that selectively retrieved visually grounded commonsense knowledge is beneficial for reasoning-intensive questions. Math (\textit{MA}) and Coarse Perception (\textit{CP}) also improved from \textit{M~ (Full)}, showing that the router mitigates degradation from fixed knowledge routes.
These results indicate that external commonsense knowledge utility is heterogeneous across inputs and question categories. Selective routing is more robust, retaining beneficial routes when estimated utility is positive and defaulting to the no-knowledge route when augmentation is unhelpful.
When applied to the VCR and VQAv2 datasets using 5-fold CV, routed MMCOMET outperforms the baseline by up to +0.0010 and +0.0114, respectively. More details for the router may be found in Appendix \ref{app:router}.

\vspace{-0.8em}
\section{Conclusion}
\vspace{-0.5em}
We introduced MMCOMET, a large-scale multimodal commonsense knowledge graph that grounds physical, social, and eventive knowledge with representative visual evidence. By extending ATOMIC2020 through a scalable hybrid alignment pipeline, MMCOMET provides nearly one million multimodal commonsense tuples for retrieval-augmented vision-language reasoning and generation. Experiments across diverse downstream tasks show that visually grounded commonsense consistently improves over text-only and LLM-generated knowledge, particularly in reasoning-intensive settings. We further propose an adaptive knowledge router for heterogeneous knowledge requirements, enabling dynamic route selection with a no-knowledge fallback when no benefit from commonsense knowledge is expected. Overall, MMCOMET establishes a reusable resource and benchmark for studying grounded commonsense integration in multimodal systems.

\vspace{-0.5em}
\section*{AI use statement}
In this work, generative AI tools were used only to edit the manuscript for grammar and readability. We have not used generative AI tools for any of the required-disclosure tasks. All such work is the authors' own. All AI-assisted edits were reviewed by authors. We take full responsibility for the final content of this work, including text, claims or artifacts.

\section*{Ethics Statement}
Further details including licensing, risk and institutional review are provided in Appendix~\ref{appendix:ethics}

\section*{Acknowledgement}
We thank Prof. Eduard Hovy and Prof. Sangwook Yi for their support of this research. This research was supported by the Institute of Information \& communications Technology Planning \& Evaluation (IITP) grant funded by the Korea government (MSIT) (No.RS-2026-25519206, Development of human-like intelligent generative agents based on interactive multimodal reverse prompting), and IITP grant funded by the Korea government (MSIT) (No.RS-2025-02217259, Development of self-evolving AI bias detection-correction-explain platform based on international multidisciplinary governance).

\bibliography{iclr2027_conference}
\bibliographystyle{iclr2027_conference}

\appendix



\clearpage

\clearpage

\begin{table}[!t]
\centering
\footnotesize
\setlength{\tabcolsep}{5pt}
\renewcommand{\arraystretch}{1.1}
\vspace{-5em}
\begin{tabularx}{\textwidth}{
  >{\raggedright\arraybackslash}p{2.5cm}
  >{\raggedright\arraybackslash}p{3cm}
  >{\centering\arraybackslash}p{2.5cm}
  >{\raggedright\arraybackslash}X
}
\toprule
\textbf{Category} & \textbf{Subcategory} & \textbf{Location} & \textbf{Key Items (Purpose)} \\
\midrule
& \multicolumn{3}{c}{\small{\textit{\textbf{How is MMCOMET constructed?}}}} \\[-2pt]
\cmidrule(l){2-4}\cmidrule(l){2-4}\\[-12pt]
\multirow{4}{2.5cm}{\small{\textbf{Construction}}}
& Base Commonsense Topology
& Sec.~\ref{subsec:base_topology}
& Table~\ref{tab:comparisontable} (MMKG Comparison), Fig.~\ref{fig:arch} (Construction Pipeline Overview) \\
\cmidrule(l){2-4}
& Construction Pipeline
& Sec.~\ref{subsec:hybrid_alignment}, Sec.~\ref{subsec:postprocessing}
& - \\
\cmidrule(l){2-4}
& Graph Statistics
& App.~\ref{appendix: rel_examples}, App.~\ref{appendix:webretrieval_relation}, App.~\ref{appendix: graph_stats}
& Table~\ref{tab: relation_description} (Relation Definitions), Table~\ref{relation_stats_table} (Web-based retrieval Relations), Table~\ref{tab: graph_stats} (Graph Statistics) \\
\midrule
& \multicolumn{3}{c}{\small{\textit{\textbf{Is MMCOMET visually well-grounded?}}}} \\[-2pt]
\cmidrule(l){2-4}\cmidrule(l){2-4}\\[-12pt]
\multirow{2}{2.5cm}{\small{\textbf{Internal Analysis}}}
& Annotation Filtering
& Sec.~\ref{sec:intrinsic_eval}, App.~\ref{appendix:human_eval_info_intrinsic}
& Table~\ref{tab: human_eval} (Full/Partial Match Rates by Relation) \\
\cmidrule(l){2-4}
& Concreteness \& Alignment Analysis
& Sec.~\ref{sec:intrinsic_eval},
App.~\ref{appendix:full_vs_partial},
App.~\ref{appendix:relation_types}
& Fig.~\ref{fig: full_vs_partial}, Fig.~\ref{fig: good_examples} \\
\midrule
& \multicolumn{3}{c}{\small{\textit{\textbf{Does visual grounding help downstream tasks?}}}} \\[-2pt]
\cmidrule(l){2-4}\cmidrule(l){2-4}\\[-12pt]
\multirow{6}{2.5cm}{\small{\textbf{Downstream Task}}}
& Evaluation Setup \& Reproducibility & App.~\ref{mmcomet-vst}, App.~\ref{appendix:eval_setup}  & Table~\ref{tab:hyperparams}, ~\ref{tab:dataset_stats} (Evaluation Setup), Fig~\ref{fig:mmcomet-vst} (VST Setup), Table~\ref{tab:prompt_vcr_baseline} - \ref{tab:prompt_imagecaptioning_mmcomet} (Prompts) \\
\cmidrule(l){2-4}
& Downstream Task Result & Sec.~\ref{sec:experimental_results} & Table ~\ref{tab:results_vcr} -- ~\ref{tab:results_ic} (Main Result; VCR, VQA, VST, IE, IC), Table~\ref{vst_automatic_metrics} \\
\cmidrule(l){2-4}
& Qualitative Analysis &
App.~\ref{appendix:qa_extrinsic} & Fig.~\ref{fig:full_qa_vcr} -- Fig.~\ref{fig:full_qa_ic} (Qualitative Analysis) \\
\midrule
& \multicolumn{3}{c}{\small{\textit{\textbf{Which knowledge actually drives the gains -- and how should it be selected?}}}} \\[-2pt]
\cmidrule(l){2-4}\cmidrule(l){2-4}\\[-12pt]
\multirow{6}{2.5cm}{\small{\textbf{Knowledge Router}}}
& Effect of Knowledge Source
& Sec.~\ref{sec:llm_ablation}
& Fig.~\ref{fig:ablation_llm_gen} (Result by Knowledge Source), Table~\ref{tab:vtkg_vcr_vqa} (Comparision with VTKG-C) \\
\cmidrule(l){2-4}
& Domain Coverage (Physical/Social/Eventive)
& App.~\ref{appendix:domain_coverage}
& Table~\ref{tab:domain_coverage}(VCR, VQA, IC) \\
\cmidrule(l){2-4}
& Alignment Strategy Ablation
& App.~\ref{appendix:alignment_strategy}
& Table~\ref{tab:alignment_strategy} (VCR \& Image Captioning by image source) \\
\cmidrule(l){2-4}
& Breaking Point
& App.~\ref{appendix:ablation_vcr}
& Fig.~\ref{fig:vcr_categories} (Category Breakdown), Fig.~\ref{fig:performance_gap} (Ambiguity vs.\ Accuracy),
  Fig.~\ref{fig: ambiguity_accuracy}, ~\ref{fig:performance_gap_category} (Performance Gap by Category) \\
\cmidrule(l){2-4}
& Router Experiment
& Sec.~\ref{sec:router}, App.~\ref{app:router}
& Table~\ref{tab:router_main} (Main Result), Table~\ref{tab:mmstar_breakdown} (Full Result), Alg.~\ref{alg:acur_train},~\ref{alg:acur_inference} (Algorithm)\\
\midrule
& \multicolumn{3}{c}{\small{\textit{What ethical safeguards are in place?}}} \\[-2pt]
\cmidrule(l){2-4}\cmidrule(l){2-4}\\[-12pt]
\multirow{4}{2.5cm}{\small{\textbf{Ethical Considerations}}}
& Released Artifacts \& Licensing
& App.~\ref{appendix:ethics}
& - \\
\cmidrule(l){2-4}
& Privacy \& Social-Inferential Risk
& App.~\ref{appendix:ethics}
& - \\
\cmidrule(l){2-4}
& Sensitive Content \& Safety Filtering
& App.~\ref{appendix:ethics}, App.~\ref{appendix:error_analysis}
& Fig.~\ref{fig: error_examples} (SafeSearch Filtering, Manual Spot-check on 10K Images) \\
\cmidrule(l){2-4}
& Bias \& Stereotyping
& App.~\ref{appendix:ethics}
& - \\
\bottomrule
\end{tabularx}
\caption{
  Structural overview of MMCOMET.
}
\label{tab:toc_v2}
\end{table}

\clearpage
\section{Overview}
\vspace{-0.8em}
This Appendix provides additional details for dataset construction, evaluation setup, extended results, and qualitative examples supporting the main paper. In particular, we organize our extended experimental analyses around three research questions:
\vspace{-0.5em}

\begin{itemize}
    \item \textbf{RQ1: Does multimodal grounding matter?} Text-only knowledge bases cannot resolve which of several plausible commonsense associations applies to a specific visual scene, so we test whether grounding commonsense in images resolves this ambiguity. We compare MMCOMET against text-only knowledge (TextKG), LLM-generated knowledge, and an existing multimodal commonsense knowledge graph (VTKG-C) in Sec.~\ref{sec:llm_ablation} and App.~\ref{appendix:llm_gen_figure}. MMCOMET outperforms all three alternatives in most settings, and the margin is largest on reasoning-heavy tasks. This suggests that visual grounding supplies information that text-only, LLM-generated, or single-domian knowledge does not reliably recover.
    \item \textbf{RQ2: Which components of MMCOMET matter?} MMCOMET's contribution combines two design choices: relation coverage beyond the physical domain, and a hybrid alignment strategy for linking phrases to images. We ablate each in Appendix~\ref{appendix:domain_coverage} and  Appendix~\ref{appendix:alignment_strategy}. The domains that matter are task-dependent. Removing social and eventive coverage disproportionately hurts VCR, while replacing the hybrid alignment with a single-signal alternative degrades retrieval precision, confirming both components are necessary. Broad coverage therefore matters more than any single task-specific optimum
    \item \textbf{RQ3: Which knowledge actually drives the gains -- and how should it be selected?} If MMCOMET's benefit reflects commonsense reasoning rather than a uniform boost, it should concentrate where reasoning is required. We analyze gains by sample ambiguity on VCR (App.~\ref{appendix:ablation_vcr}) and by question category on MMStar (Sec.~\ref{sec:router}, App.~\ref{app:router}). The gap between MMCOMET and TextKG widens with ambiguity, and on MMStar full injection helps Logical Reasoning for larger models but hurts perception-heavy categories and overall accuracy. This motivates an adaptive utility router that selects a knowledge route or falls back to baseline, which recovers or slightly exceeds the baseline. We therefore view it as evidence for selective knowledge use rather than a large accuracy gain.
\end{itemize}

\vspace{-1.2em}
\section{Ethics Statement}
\label{appendix:ethics}
\vspace{-0.8em}
This work proposes a resource that grounds text-only commonsense knowledge with visual evidence, most of which is retrieved via web search. We acknowledge that this raises additional concerns, which we attempt to address in this section.
\vspace{-1em}
\paragraph{Limitations}
First, our retrieval pipeline relies on CLIP ViT-B/32 embeddings, which, while comparatively dated, reflect a deliberate scalability/efficiency trade-off at the scale of our benchmark ($\sim$3.9M images); adopting a stronger encoder is left for future work. Second, MMCOMET is constructed from web-crawled images, which raises ethical considerations around data provenance, consent, and the potential presence of sensitive or inappropriate content that automated crawling cannot fully filter out; we discuss our mitigation efforts (e.g., filtering pipelines) in the ethics statement, and note that source URL expiration over time may also affect long-term reproducibility, which we mitigate by releasing the full precomputed image feature matrix alongside the benchmark. Third, we fix the number of retrieved phrases $k$ to a single, empirically stable value across all tasks rather than tuning it per task to keep the framework simple and broadly applicable; task-specific retrieval depths (e.g., for VQA, VCR, or captioning) may yield further gains, which we leave to future work.
\vspace{-1.5em}
\paragraph{Released Artifacts and Licensing.} For all images, we release only the final visual representations used to construct the resource, along with the complete triplet-image association and embedding-extraction metadata. We do not redistribute raw images, whether web-sourced or from other open-source datasets. For web-retrieved images, we further release provenance metadata and licensing details. We do not include URLs to respect copyright and privacy constraints and to prevent potential reverse identification.
\vspace{-1em}
\paragraph{Privacy and Social-Inferential Risk.} We acknowledge that some images used may contain identifiable individuals that has been systematically paired with inferred attributes, intents, and emotional states. We attempt to mitigate this by not releasing the full raw images however, we don't have full control for images from open-source datasets. Instead, we provide a clear take down mechanism for any image or triplet-image association upon request to the corresponding author or through the released repository.
\vspace{-1em}
\paragraph{Sensitive Content and Safety Filtering.} For all web-retrieved images, we used the Google Custom Search API with active SafeSearch enabled. This provides us with robust safety filtering against sensitive and explicit content as illustrated in Figure \ref{fig: error_examples} and discussed in Appendix~\ref{appendix:error_analysis}. However, we acknowledge that enabling this option is not 100\% perfect. As such, we conducted random spot-checking and manual evaluation on 10k images and 1000 phrases.
\vspace{-1em}
\paragraph{Bias and Stereotyping.} MMCOMET uses naturally occurring web imagery to preserve the diversity of real-world visual contexts rather than relying only on curated benchmark images. At the same time, we acknowledge that real-world web data can reproduce existing social and demographic biases, including occupational, gender, and racial stereotypes.
\vspace{-1em}
\paragraph{Institutional Review Board Statement.}
The study was approved by the Institutional Review Board (Exemption Number: PIRB-2025-E020, Date: June 9, 2025).

\vspace{-0.5em}
\section{Multimodal Commonsense Enhanced Visual Storytelling} \label{mmcomet-vst}
\begin{figure*}[!t]
  \centering
  \includegraphics[width=0.7\linewidth]{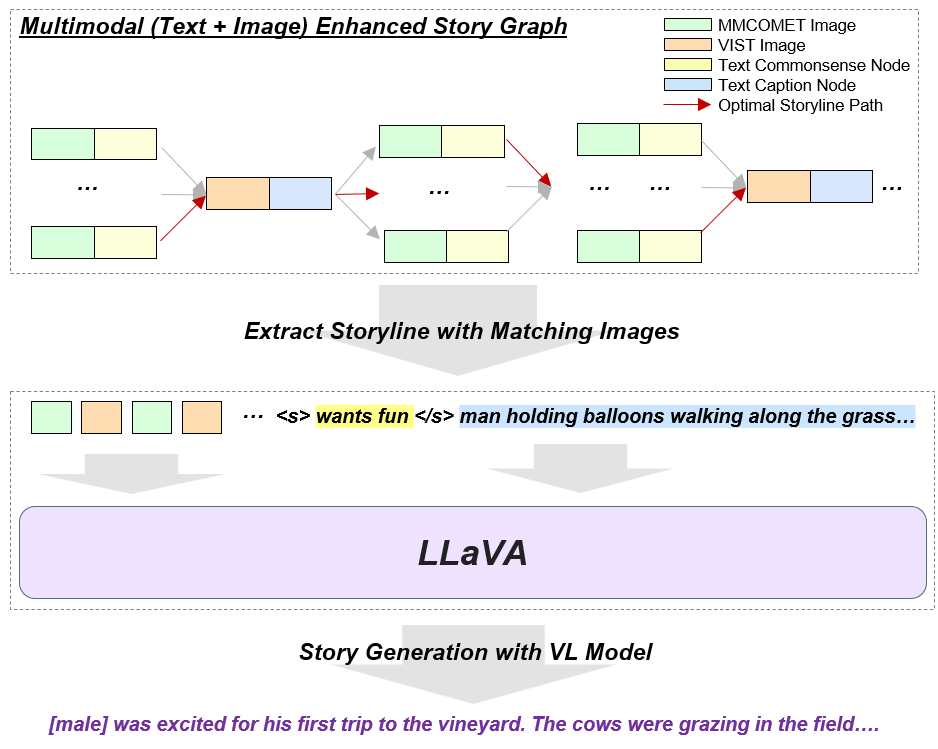}
  \caption{SCO-VIST framework enhanced with MMCOMET.}
  \label{fig:mmcomet-vst}
\end{figure*}

Figure \ref{fig:mmcomet-vst} depicts how SCO-VIST's storygraph is enhanced with information from MMCOMET. In SCO-VIST, each node in the graph only has a textual modality (e.g. a caption description or commonsense phrase). We incorporate image features into the story graph by using the generated phrase to query MMCOMET to find the matching image. For simplicity, if the phrase is not found in MMCOMET, we delete the node from the story graph. The CLIP features of the top-found image are then concatenated with the textual features for the commonsense nodes in the story graph, while for the caption nodes, we directly use the image features from the image sequence prompt. The story graph edge weights are then recalculated using the cosine similarity between adjacent nodes. The storyline is then extracted from the graph using a shortest-path search algorithm. The textual storyline, along with the MMCOMET and VIST images, is finally fed into a vision-language model (LLaVA) for story generation. Additionally, two integration setups are tested. LLaVA+MMCOMET$^{S}$ enriches the story graph with MMCOMET image embeddings, and the input into the LLaVA story decoder is the storyline obtained from the multimodal story graph + 5 VIST images. LLaVA+MMCOMET$^{S+I}$ additionally feeds the MMCOMET images, the enriched storyline, and the 5 VIST images.

\vspace{-1em}
\section{Multimodal Commonsense Graph Analysis}
\label{appendix:multimodal_graph}
\vspace{-0.5em}
\subsection{Comet-ATOMIC2020 Relation Types} \label{appendix: rel_examples}
Table \ref{tab: relation_description} shows the different relations and their definitions, as well as corresponding examples in the original Comet-ATOMIC2020 graph. Green, orange and blue highlighting indicates social, event and physical
relations respectively. Note that for simplicity, in this paper, we refer to \texttt{xEffect} and \texttt{oEffect} as \texttt{Effect}, \texttt{xReact} and \texttt{oReact} as \texttt{React}, and \texttt{xWant} and \texttt{oWant} as \texttt{Want}. The \texttt{x} prepended in front of the relation name is also removed for consistency e.g. \texttt{xNeed} is referred to as \texttt{Need}.

\begin{table*}[!t]
  \caption{Different Relations in Comet-ATOMIC2020 and corresponding examples. Examples and definitions taken from the official Comet-ATOMIC2020 paper~\citep{hwang2021comet}.}
  \resizebox{1\textwidth}{!}{  
  \begin{tabular}{@{}l|l|l}
    \hline
    \textbf{Relation} & \textbf{Description} & \textbf{Example ($<$head, \texttt{relation}, tail$>$)}\\ \hline
    ObjectUse  & describes everyday affordances or uses of objects & bread \texttt{ObjectUse} make french toast \\
    AtLocation & spatially describes the where an entity is likely to be found & bread \texttt{AtLocation} pantry \\
    MadeUpOf  & describes a part, portion or makeup of an entity & bread \texttt{MadeUpOf} dough \\
    HasProperty  & describes entities’ general characteristics & bread \texttt{HasProperty} nice to eat \\
    CapableOf  & describe abilities and capabilities of everyday living entities & baker \texttt{CapableOf} coat cake with icing \\
    Desires & desire of sentient entity & baker \texttt{Desires} quality ingredients \\
    NotDesires & non-desire of sentient entity & baker \texttt{NotDesires} bad yeast \\ \hline
    isAfter & events that can follow an event & X runs out of steam \texttt{isAfter} X exercises in the gym \\
    HasSubEvent & provides the internal structure of an event & X runs out of steam \texttt{HasSubEvent} become tired \\
    isBefore & events that can precede an event & X runs out of steam \texttt{isBefore} X hits the showers \\
    HinderedBy & hindrances that obstruct the natural path of an event & X runs out of steam \texttt{HinderedBy} drinks too much coffee \\
    Causes & causal relation between two events or entities & X runs out of steam \texttt{Causes} takes a break \\ 
    xReason & provides a post-fact explanation of the cause of an event & X runs out of steam \texttt{xReason}  did not eat breakfast \\ \hline
    xNeed & describes a precondition for an agent to achieve the event & X runs out of steam \texttt{xNeed}  do something tiring \\
    xAttr & describes personas or attributes perceived by others given an event &  X runs out of steam \texttt{xAttr} old  \\
    xEffect & actions that happen to agent X that may occur after the event & X runs out of steam \texttt{xEffect}  drinks some water \\
    oEffect & actions that happen to agent Y that may occur after the event & X votes for Y \texttt{oEffect} receives praise \\
    xReact & emotional reactions of agent X in an event & X runs out of steam \texttt{xReact} tired \\
    oReact & emotional reactions of agent Y in an event & X votes for Y \texttt{oReact} grateful \\
    xWant & agent X's postcondition desires after an event & X runs out of steam \texttt{xWant} to get some energy \\
    oWant & agent Y's postcondition desires after an event & X votes for Y \texttt{oWant} thank X \\
    xIntent &  defines the likely intent of an agent & X votes for Y \texttt{xIntent} to give support \\
    \hline
  \end{tabular}} 
  \label{tab: relation_description} 
\end{table*}

\vspace{-0.5em}
\subsection{Web Retrieval Relation Type Analysis}
\label{appendix:webretrieval_relation}
Table~\ref{relation_stats_table} reports the proportion of head and tail phrases routed to web-based retrieval under the concreteness threshold. Consistent with their lower concreteness scores, social and eventive relations require substantially more web searches than physical relations. 
An interesting asymmetry emerges within certain physical relations such as \texttt{HasProperty} and \texttt{Desires/NotDesires}. While head phrases typically contain tangible noun entities and therefore exhibit high concreteness, tail phrases often describe abstract properties or mental states. As a result, tail phrases in these relations show a disproportionately high reliance on web-based grounding. The analysis confirms a strong correlation between relation-level concreteness and retrieval routing behavior. These findings empirically justify the concreteness-aware hybrid alignment strategy, as a uniform similarity-based approach would inadequately ground abstract social and eventive commonsense expressions.

\begin{table*}[!ht]
\small
  \centering
  \tiny  
  \caption{\label{relation_stats_table}
    The left sub-table shows the average concreteness scores for each commonsense relation (sample size in brackets) in ascending order. The second and third sub-tables show the proportion of heads and tail phrases across each relation that required web search. Green, orange and blue highlighting indicate social, event and physical relations, respectively.
  }
  \resizebox{1\textwidth}{!}{\begin{tabular}{p{2cm}c||p{1.5cm}c||p{1.5cm}c}
    \hline
    \textbf{Relation}  & \textbf{Avg. Concreteness} & \textbf{Relation} & \textbf{\% Heads Searched} & \textbf{Relation} & \textbf{\% Tails Searched}\\
    \hline
    \cellcolor{green!15}React (94,228) & 2.95 & \cellcolor{cyan!15}NotDesires & 2.82 & \cellcolor{cyan!15}AtLocation & 25.52 \\
    \cellcolor{green!15}Attr (115,815) & 2.98 & \cellcolor{cyan!15}Desires & 2.85 & \cellcolor{cyan!15}MadeUpOf & 36.47 \\
    \cellcolor{orange!15}Causes (376) & 3.29 & \cellcolor{cyan!15}AtLocation & 18.78 & \cellcolor{orange!15}isAfter & 48.98 \\
    \cellcolor{green!15}Intent (49,312) & 3.40 & \cellcolor{cyan!15}MadeUpOf & 20.03 & \cellcolor{orange!15}isBefore & 53.50 \\
    \cellcolor{green!15}Effect (113,494) & 3.53 & \cellcolor{cyan!15}HasProperty & 22.77 & \cellcolor{cyan!15}CapableOf & 59.55 \\
    \cellcolor{orange!15}Reason (334) & 3.55 & \cellcolor{cyan!15}CapableOf & 23.98 & \cellcolor{orange!15}HinderedBy  & 64.72 \\ 
    \cellcolor{orange!15}HasSubEvent (12,845) & 3.58 & \cellcolor{cyan!15}ObjectUse & 25.26 & \cellcolor{cyan!15}ObjectUse & 67.65 \\ 
    \cellcolor{green!15}Want (152,989) & 3.58 & \cellcolor{orange!15}Reason & 50.90 & \cellcolor{cyan!15}HasProperty & 71.46 \\ 
    \cellcolor{green!15}Need (89,391) & 3.65 & \cellcolor{green!15}Intent & 51.76 & \cellcolor{orange!15}HasSubEvent & 73.03 \\ 
    \cellcolor{orange!15}HinderedBy (106,647) & 3.80 & \cellcolor{green!15}Need & 54.11 & \cellcolor{cyan!15}NotDesires & 74.03 \\ 
    \cellcolor{cyan!15}HasProperty (5,617) & 3.80 & \cellcolor{green!15}React & 54.85 & \cellcolor{green!15}Need & 75.94 \\ 
    \cellcolor{cyan!15}Desires (2,737) & 3.88 & \cellcolor{green!15}Effect & 54.85 & \cellcolor{orange!15}Reason & 76.35 \\ 
    \cellcolor{cyan!15}ObjectUse (165,590) & 3.90 & \cellcolor{orange!15}isAfter & 55.96 & \cellcolor{green!15}Effect & 76.54 \\ 
    \cellcolor{orange!15}isBefore (23,208) & 3.91 & \cellcolor{orange!15}isBefore & 56.07 & \cellcolor{orange!15}Causes & 77.93 \\ 
    \cellcolor{cyan!15}NotDesires (2,838) & 3.91 & \cellcolor{green!15}Want & 56.15 & \cellcolor{cyan!15}Desires & 79.65 \\ 
    \cellcolor{orange!15}isAfter (22,453) & 3.94 & \cellcolor{orange!15}HinderedBy & 56.19 & \cellcolor{green!15}Want & 80.92 \\ 
    \cellcolor{cyan!15}CapableOf (7,968) & 3.96 & \cellcolor{green!15}Attr & 57.13 & \cellcolor{green!15}Intent & 90.35 \\ 
    \cellcolor{cyan!15}MadeUpOf (3,345) & 4.08 & \cellcolor{orange!15}HasSubEvent & 60.19 & \cellcolor{green!15}Attr & 98.82 \\ 
    \cellcolor{cyan!15}AtLocation (20,221) & 4.15 & \cellcolor{orange!15}Causes & 65.96 & \cellcolor{green!15}React & 99.48\\
    \hline
  \end{tabular}}
\end{table*}

\vspace{-0.5em}
\subsection{Graph Statistics} \label{appendix: graph_stats}
Table \ref{tab: graph_stats} shows the number of unique nodes, total edges, unique web searched images, unique images acquired from open-source datasets and total unique images for each relation, and across all relations. 

\subsection{Error Analysis}
\label{appendix:error_analysis}

\begin{figure*}[!h]
  \centering
  \includegraphics[width=0.8\linewidth]{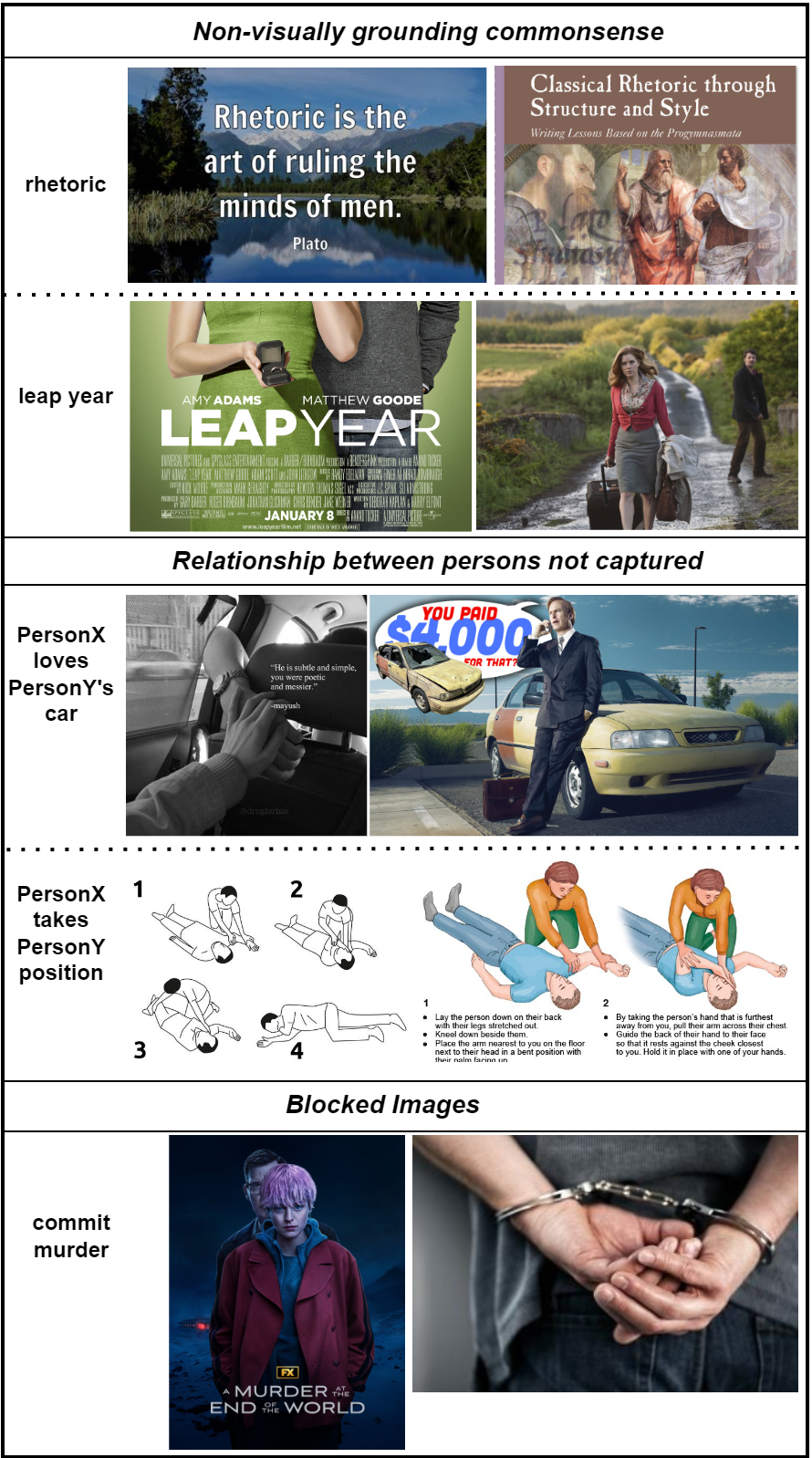}
  \caption{Examples of retrieved images for 3 different error cases found during the image collection stage.}
  \label{fig: error_examples}
\end{figure*}

We analyse common errors that occurred in the process of finding matching images for the commonsense phrase. The 3 main types of errors and visual examples are depicted in Figure \ref{fig: error_examples}. One challenging issue that remains is finding matching images for non-visually grounding commonsense. For instance, the first 2 commonsense examples in Figure \ref{fig: error_examples}, `\textit{rhetoric}' and `\textit{leap year}' are very abstract terms and hence, difficult to visually ground even when employing web search. Consequently, the images obtained usually contains text or quotes as with the case for the matched images found for `\textit{rhetoric}'. In other cases, the retrieved images may sometimes be related to movies or media where the title of the media is similar to the commonsense phrase, such as with the case of `\textit{leap year},' which seems to have retrieved images related to a movie that has the same name as the commonsense phrase. 
The second issue found was that sometimes the relationship between people and concepts in the commonsense phrase is not sufficiently captured by the image. This typically occurs when the phrase mentions two people - `\textit{PersonX}'and `\textit{PersonY}'. For instance, the matching images found from `\textit{PersonX loves PersonY's car}' either solely focus on the idea of `love' as in the left image or the concept of `car' as in the image on the right. However, the complex relationship between the first concept of a `person loving' and the second concept, `another person's car', is much more difficult to capture. Finally, there are cases where the commonsense phrase mentions violent or inappropriate topics. Inevitably, these images are most likely blocked by Google's web search engine, such that the images obtained may not fully match the search phrase. For example, for the commonsense phrase `\textit{commit murder}', we again obtained images related to movies/television shows as shown in the left image. At best, we tend to obtain images that are partial matches, such as in the image on the right, where the action is not explicitly shown but is implied. 

\vspace{-0.5em}
\section{Human Evaluation Information}
\label{appendix:human_eval_info}
\subsection{Intrinsic Evaluation}
\label{appendix:human_eval_info_intrinsic}
We conducted our annotation process with three annotators: a PhD student, a Postdoctoral researcher, and an Associate Professor. These annotators were recruited based on their expertise in the relevant research to ensure high-quality annotations. Our annotators come from diverse geographical backgrounds—Asia, Australia, and Europe—reflecting a broad perspective in the annotation process.
The human evaluation was conducted on a voluntary basis, and no monetary compensation was provided. Before participation, we provided clear instructions explaining the purpose of the evaluation and how the collected data would be used. The annotators were informed that their feedback would contribute to the assessment and refinement of our system's performance, and they participated with full awareness and consent.

\begin{table*}[!h]
  \centering  
  \footnotesize
  \caption{Graph statistics for MMCOMET showing total unique nodes, edges, web-searched images, open-source dataset images and total images across the different relations.}
  \begin{tabular}{@{}c||ccccc}  
    \hline
     \textbf{Relation} & \textbf{\# Nodes} & \textbf{\# Edges} & \textbf{\# WSE Images} & \textbf{\# OID Images} & \textbf{\# Total Images} \\ \hline
    ObjectUse & 137,144 & 165,590 & 767,124 & 242,211 & 1,009,335 \\
    AtLocation & 11,216 & 20,221 & 35,204 & 62,730 & 97,934 \\
    MadeUpOf & 3,464 & 3,345 & 11,047 & 24,651 & 35,698 \\ 
    HasProperty  & 6,123 & 5,617 & 32,233 & 35,738 & 67,971 \\
    CapableOf  & 8,872 & 7,968 & 45,005 & 44,675 & 89,680 \\
    Desires  & 2,832 & 2,737 & 18,750  & 14,911 & 33,661 \\
    NotDesires & 2,991 & 2,838 & 18,312  & 15,602 & 33,914 \\
    \hline
    isAfter & 36,181 & 22,453 & 162,312 & 97,386 & 259,698 \\
    HasSubEvent & 10,670 & 12,845 & 67,522 & 42,541 & 110,063 \\
    isBefore & 37,482 & 23,208 & 175,957 & 96,981 & 272,938 \\ 
    HinderedBy  & 103,844 & 106,647 & 588,983 & 164,975 & 753,958 \\
    Causes  & 531 & 376 & 3,299 & 3,471 & 6,770 \\
    Reason  & 441 & 334 & 2,622 & 2,888 & 5,510 \\
    \hline
    Need & 80,575 & 89,391 & 490,672 & 136,668 & 627,340 \\
    Attr & 29,650 & 115,815 & 175,477 & 73,072 & 248,549 \\
    Effect & 90,983 & 113,494 & 567,801 & 134,982 & 702,783 \\ 
    React  & 35,537 & 94,228 & 228,856 & 77,635 & 306,491\\
    Want  & 107,268 & 152,989 & 689,798 & 146,314 & 836,112\\
    Intent  & 45,771 & 49,312 & 292,494 & 92,549 & 385,043 \\
    \hline
   \textbf{All} & 571,404 & 989,408 & 3,476,098 & 435,695 & 3,911,793 \\
    \hline
  \end{tabular} \label{tab: graph_stats} 
\end{table*}

\subsection{Extrinsic Evaluation}
\label{appendix:human_eval_info_extrinsic}
We conducted our annotation process with eleven annotators, including one undergraduate student, two graduated assistants, four Master's students, three PhD students, and one Postdoctoral researcher. These annotators were recruited based on their familiarity with the task and relevant research experience to ensure high-quality annotations. 
All annotators are from Asia, providing a consistent cultural background for the evaluation process. The group consisted of five male and six female annotators, ensuring gender balance.
Human evaluation was conducted on a voluntary basis and no monetary compensation was provided. Before participation, we provided clear instructions explaining the purpose of the evaluation and how the collected data would be used. The annotators were informed that their feedback would contribute to the assessment and refinement of our system's performance, and they participated with full awareness and consent.

Each annotator evaluated 80 samples drawn from four tasks that we evaluated (VCR, VQA, Image Editing, Image Captioning), with 20 samples per task. For each sample, annotators were presented with the input image, the model outputs under each setting, and the retrieved commonsense knowledge phrases provided to model. They rated two statements on a 4-point Likert scale (1 = Strongly Disagree, 4 = Strongly Agree), (1) \textit{"Is the retrieved commonsense knowledge relevant to the image and input?"} and (2) \textit{"Does the knowledge help improve reasoning or inference for this task?"}. 

The average scores across all tasks were 3.1 and 2.9 for Q1 and Q2, respectively, both exceeding the neutral midpoint of 2.5. These results suggest that annotators generally found the retrieved commonsense knowledge to be relevant and beneficial, even when evaluated across both success and fail cases. Notably, even in cases where MMCOMET did not outperform the Baseline on automatic metrics, annotators still perceived the retrieved knowledge as helpful for reasoning, highlighting a potential gap between automatic evaluation metrics and human judgement.

\begin{wraptable}[18]{r}{0.35\columnwidth}
\vspace{-5pt}
\centering
\setlength{\tabcolsep}{4pt}
\caption{Relevant hyperparameters.}
\label{tab:hyperparams}
\small
\begin{tabular}{lc}
\toprule
\textbf{Hyperparameter} & \textbf{Value} \\
\midrule
\multicolumn{2}{c}{\textit{Common}} \\
\midrule
CLIP backbone & ViT-B/32 \\
Candidate images ($n$) & 15 \\
Retrieved phrases ($k$) & 5 \\
Random seed & 42 \\
\midrule
\multicolumn{2}{c}{\textit{Max Tokens}}\\
\midrule
Max Tokens - VCR & 10 \\
Max Tokens - VQA & 10 \\
Max Tokens - Captioning & 64 \\
\midrule
\multicolumn{2}{c}{\textit{Image Editing}}\\
\midrule
Inference steps & 50 \\
CFG scale & 4.0 \\
Retrieved phrases ($k$) & 1 \\
\bottomrule
\end{tabular}

\end{wraptable}

\section{Evaluation Setup}
\label{appendix:eval_setup}
\begin{table*}[t]
\centering
\caption{Task dataset statistics, metrics, and models used.}
\label{tab:dataset_stats}
\scriptsize
\begin{tabular}{
>{\centering\arraybackslash}p{0.15\textwidth}
>{\centering\arraybackslash}p{0.1\textwidth}
>{\centering\arraybackslash}p{0.1\textwidth}
>{\centering\arraybackslash}p{0.25\textwidth}
>{\centering\arraybackslash}p{0.2\textwidth}
}
\toprule
\textbf{Task} & \textbf{Dataset} & \textbf{\#Items} & \textbf{Metric} & \textbf{Models}\\
\midrule
Visual Commonsense Reasoning
  & VCR (val) & 26,534
  & \textbf{Accuracy} & Qwen3-VL-8B-Instruct, Qwen3-VL-32B-Instruct, InternVL3-38B-hf \\
\midrule
Visual QA
  & VQAv2 (val) & 214,354
  & \textbf{Accuracy} \textit{(VQA-style soft scoring; a predicted answer
    receives partial credit based on agreement with human annotations)} & Qwen3-VL-8B-Instruct, gemma-4-31B-it  \\
\midrule
Image Editing
  & PIE-Bench & 700
  & \textbf{PSNR} \textit{(pixel-level distortion)},
    \textbf{SSIM} \textit{(structural similarity)},
    \textbf{LPIPS} \textit{(perceptual similarity)} 
  & Qwen-Image-Edit \\
\midrule
Image Captioning
  & Flickr30K (1K Test set) & 1,000
  & \textbf{BLEU} \textit{(n-gram precision)},
    \textbf{CIDEr} \textit{(consensus-based similarity)},
    \textbf{ROUGE-L} \textit{(longest common subsequence)} 
  & Qwen3-VL-8B-Instruct, Qwen3-VL-32B-Instruct, gemma-4-31B-it \\
\midrule
Visual Story Telling
    & VIST & Train: 8,031 Val: 998 Test: 1,011 & \textbf{RoViST} \textit{(visual grounding, language coherence, repetition)}, 
    \textbf{Perplexity} \textit{(NLG metric)}, 
    \textbf{UNION} \textit{(NLG metric)}, 
    \textbf{SPICE} \textit{(semantic similarity)}, 
    \textbf{BLEURT} \textit{(semantic similarity)},
    \textbf{MoverScore} \textit{(semantic similarity)}  & SCOVIST+LLAVA\\
\bottomrule
\end{tabular}
\end{table*}

Table~\ref{tab:dataset_stats} shows the dataset statistics, metric descriptions, and primary models tested for each task. We provide the relevant hyperparameters and details in Table~\ref{tab:hyperparams}. 
For all tasks except Visual Story Telling, we compare three conditions: \textbf{Baseline} (no knowledge-injected), \textbf{TextKG} (text-based knowledge injection), and \textbf{MMCOMET} (multimodal knowledge injection). All experiments were conducted with single H100 GPU with 94GB GPU memory.

For the knowledge retrieval we tried several methods including (1) \textbf{L2 retrieval}, which retrieves the top-$n$ visually similar images and re-ranks associated phrases by $s = \|E_\text{img} - E_\text{phrase}\|_2 + \|E_\text{query} - E_\text{phrase}\|_2$, (2) \textbf{FAISS-based direct search}, which retrieves phrases directly from the full phrase index using a combined embedding $E_\text{combined} = (E_\text{img} + E_\text{query}) / 2$, (3) \textbf{Weighted scoring}, which computes $s = w_{i2i} \cdot \text{sim}(E_\text{img}, E_\text{img}^\text{KG}) + w_{i2t} \cdot \text{sim}(E_\text{img}, E_\text{phrase}) + w_{t2t} \cdot \text{sim}(E_\text{query}, E_\text{phrase})$  and (4) \textbf{MMR}, which promotes diversity by $\text{MMR} = \lambda \cdot s_\text{rel} - (1-\lambda) \cdot \max_{p_j \in S} \text{sim} (p_i, p_j)$, where $S$ is the set of selected phrases. 
As performance differences across methods were negligible, making the choice of retrieval method a minor implementation detail rather than a critical design choice. Consequently, we adopt L2 retrieval for all experiments due to its simplicity. 

Regarding knowledge injection, \textbf{TextKG} inserts $k$ KG triplets, which are combinations of 2 phrases, retrieved via text query into the prompt. \textbf{MMCOMET} inserts $k$ commonsense phrases from both head and tail entities, yielding $2k$ phrases per sample. We empirically set $k=5$, as it provided consistently strong performance across tasks (Image Editing shows consistently strong when $k=1$.). Exploring task-specific optimal values of $k$ remains a direction for future work.

\paragraph{VCR} VCR\footnote{https://visualcommonsense.com/}~\citep{mmcomet_vcr} is a visual commonsense reasoning benchmark that originally requires models to select a correct answer given a question (Q$\rightarrow$ A) or to select a rationale given a question-answer pair (QA$\rightarrow$ R). In this work, we intentionally modify the task setup to require models to select a rationale given only the question (Q$\rightarrow$ R), without access to the ground-truth answer. This modification is designed to prevent shortcut learning, where a model could otherwise exploit the provided answer to trivially infer the rationale. The prompt template used for this task is shown in Table~\ref{tab:prompt_vcr_baseline} - \ref{tab:prompt_vcr_mmcomet}.

\paragraph{VQA} VQAv2\footnote{https://huggingface.co/datasets/HuggingFaceM4/VQAv2}~\citep{vqa_v2_2017} is a multimodal reasoning benchmark that asks open-ended questions about images. The questions require an understanding of vision, language and commonsense knowledge to answer. The prompt template used for this task is shown in Table~\ref{tab:prompt_vqa_baseline} -- \ref{tab:prompt_vqa_mmcomet}.

\paragraph{Image Captioning} Flickr30K\footnote{https://huggingface.co/datasets/nlphuji/flickr30k}~\citep{young2014image} is a widely used image captioning benchmark. We evaluate on the Karpathy test split~\citep{karpathy2015deep}, which is the standard evaluation protocol for image captioning tasks. The prompt template used for this task is shown in Table~\ref{tab:prompt_imagecaptioning_baseline} -- \ref{tab:prompt_imagecaptioning_mmcomet}.

\paragraph{Image Editing} PIE\_Bench\footnote{https://cure-lab.github.io/PnPInversion/}~\citep{mmcomet_piebench} is a benchmark for evaluating text-guided image editing models, consisting of 700 images across 10 editing categories. We evaluate edited image quality using PSNR, SSIM, and LPIPS, which measure pixel-level fidelity, perceptual structural similarity, and learned perceptual distance, respectively. The prompt template used for this task is shown in Table~\ref{tab:prompt_imageediting_baseline} and \ref{tab:prompt_imageediting_mmcomet}.

\paragraph{Visual Storytelling} 
Visual Storytelling Dataset\footnote{https://visionandlanguage.net/VIST/dataset.html}~\citep{huang2016visual} contains $\sim$81K unique images obtained from Flickr. Each album contains image sequences comprising 5 photos and 5 human-written stories, and each story usually comprises one sentence per image. The unique number of stories is $\sim$50K. About 67\% of the images also have matching human-written captions. We used the story sentences and captions to create the noun tag dictionary.


\section{Comprehensive Results: VST}
\label{appendix:add_results_vst}
Table~\ref{vst_automatic_metrics} reports automatic evaluation results on the VIST benchmark.  We compare baseline storytelling systems with variants augmented using MMCOMET-based multimodal commonsense retrieval.
Incorporating MMCOMET improves visual grounding (R-VG), coherence (R-C), and non-redundancy (R-NR). The best-performing configuration (LLaVA+MMCOMET$^{S+I}$) achieves the highest overall composite score (R+S+B+M+U = 272.0), substantially outperforming non-augmented vision-language baselines.
Notably, improvements are most pronounced in grounding-sensitive metrics (R-VG) and semantic similarity measures (BLEURT, MoverScore), indicating that multimodal commonsense contributes additional contextual cues beyond raw visual features. 

\begin{table*}[!h]
  \centering
  \caption{Automatic evaluation results of visual storytelling. MMCOMET-enhanced models (bottom rows) outperform baselines in visual grounding, coherence, and overall story quality metrics, demonstrating the benefit of multimodal commonsense knowledge
  (for \textbf{Perplexity}, lower score is better.)
}
  \resizebox{1\linewidth}{!}{
  \begin{tabular}{@{}l||ccc|c|ccccc|c|cc@{}}
    \toprule
     \textbf{Model} & \textbf{R-VG} & \textbf{R-C} & \textbf{R-NR}  & \textbf{RoViST (R)}  & \textbf{SPICE (S)} & \textbf{BLEURT (B)} & \textbf{MoverScore (M)} & \textbf{UNION (U)} & \textbf{Perplexity} & \textbf{R+S+B+M+U} & \textbf{Story L.} \\ \midrule
    SCO-VIST & 70.3 & 72.3 & 90.5 & 77.7  & \textbf{10.9} & 34.9 & \underline{56.0} & \underline{84.0} & 14.9 & 263.5 & 52.3 \\ 
    VILT & 70.2 & 76.3 & 94.2 & 80.2 & 9.3 & 34.3 & 52.1 & 83.3 & 12.1 & 259.2 & 50.1 \\
    LLaVa & 72.3 & 78.1 & \underline{96.1} & 82.2 & 9.5 & 34.1 & 53.9 & \textbf{84.2} & 10.5 & 263.9 & 51.9 \\
    +MMCOMET$^{S}$ & \underline{73.3} & \underline{78.4} & 95.8 & \underline{82.5} & 9.9 & \underline{35.2} & 54.2 & 82.6 & \textbf{10.8} & \underline{264.4} & 50.8 \\ 
    +MMCOMET$^{S+I}$ & \textbf{75.2} & \textbf{79.1} & \textbf{96.2} & \textbf{83.5} & \underline{10.2} & \textbf{36.7} & \textbf{57.4} & \textbf{84.2} & \underline{11.2} & \textbf{272.0} & 53.4 \\
    \bottomrule
  \end{tabular}
  }
  \label{vst_automatic_metrics}
\end{table*}

\section{Ablation Studies}
\label{appendix:ablation_studies}
\subsection{VCR Category and Ambiguity Analysis}
\label{appendix:ablation_vcr}
\paragraph{Breakdown Categories} Following VCR~\citep{mmcomet_vcr}, we categorise VCR questions into seven categories based on keyword patterns: Explanation (37.5\%), Activity (14.7\%), Temporal (4.9\%), Mental (5.1\%), Role (0.5\%), Scene (9.9\%), Hypothetical (4.7\%). Questions not matching any keyword pattern are classified as Others (22.8\%). As the authors note that the complete keyword patterns and code were not released, we manually inspected the questions and extended the keyword list for each category accordingly. Figure~\ref{fig:vcr_categories} shows representative examples of frequently retrieved phrases and images for each category. We excluded the Role category because it has a significantly lower percentage compared to other categories.

To further investigate when multimodal commonsense knowledge is most beneficial, we analyse model performance based on ambiguity, defined as $Ambiguity = 1\space-(p(top_1)\space - p(top_2))$, where $p(top_1)$ and $p(top_2)$ denote the predicted probabilities assigned to the top-1 and top-2 answer candidates by the baseline model, respectively. A higher ambiguity score indicates that the model assigns similar probabilities to multiple candidates, reflecting samples where the correct answer is less distinguishable, or inherently harder samples.

\begin{figure}[!h]

  \centering
\includegraphics[width=0.48\textwidth]{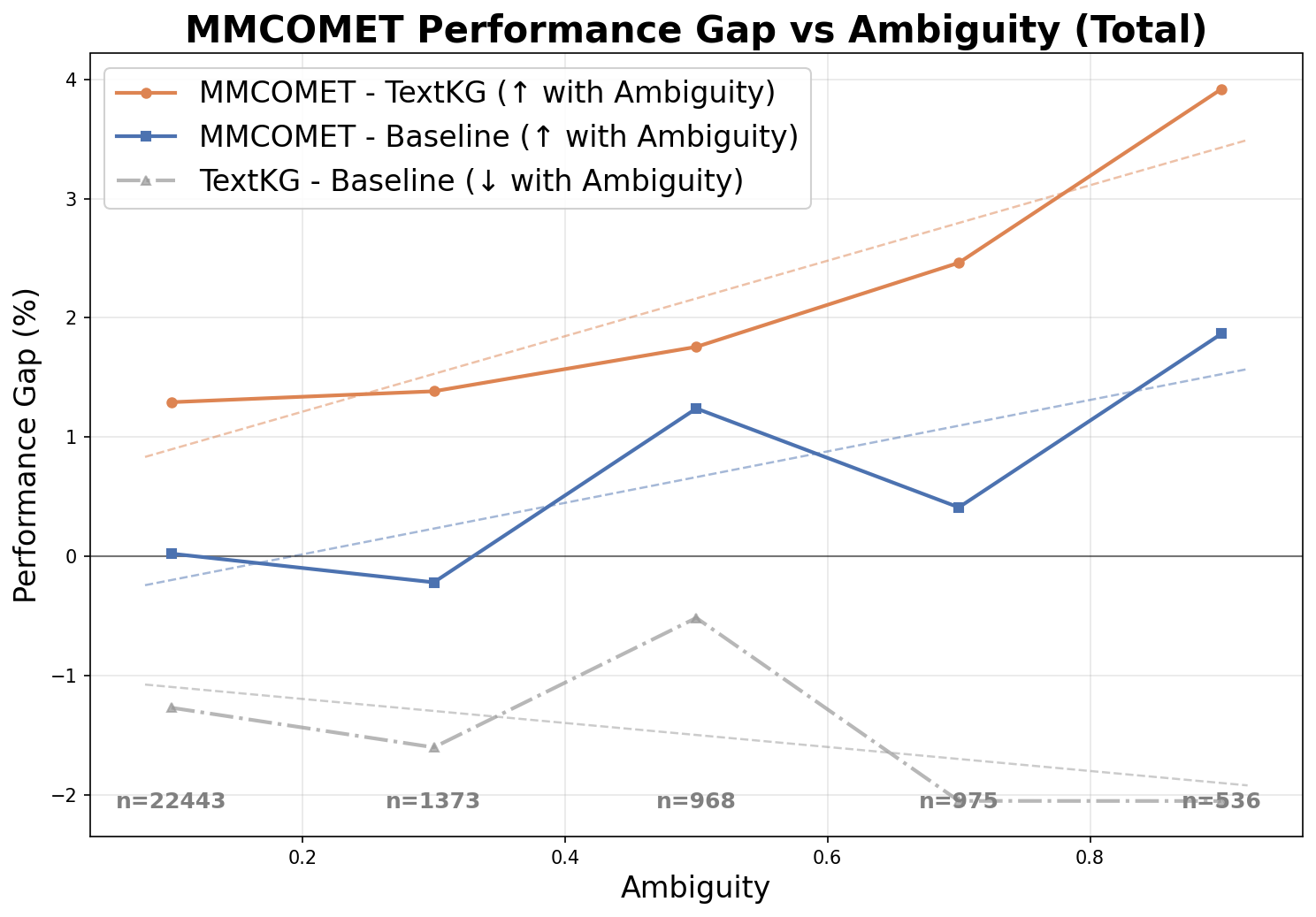}
  \caption{Ambiguity vs Performance Gap}
  \label{fig:performance_gap}
\end{figure}

\begin{figure*}[!h]
    \centering
    \includegraphics[width=0.48\textwidth]{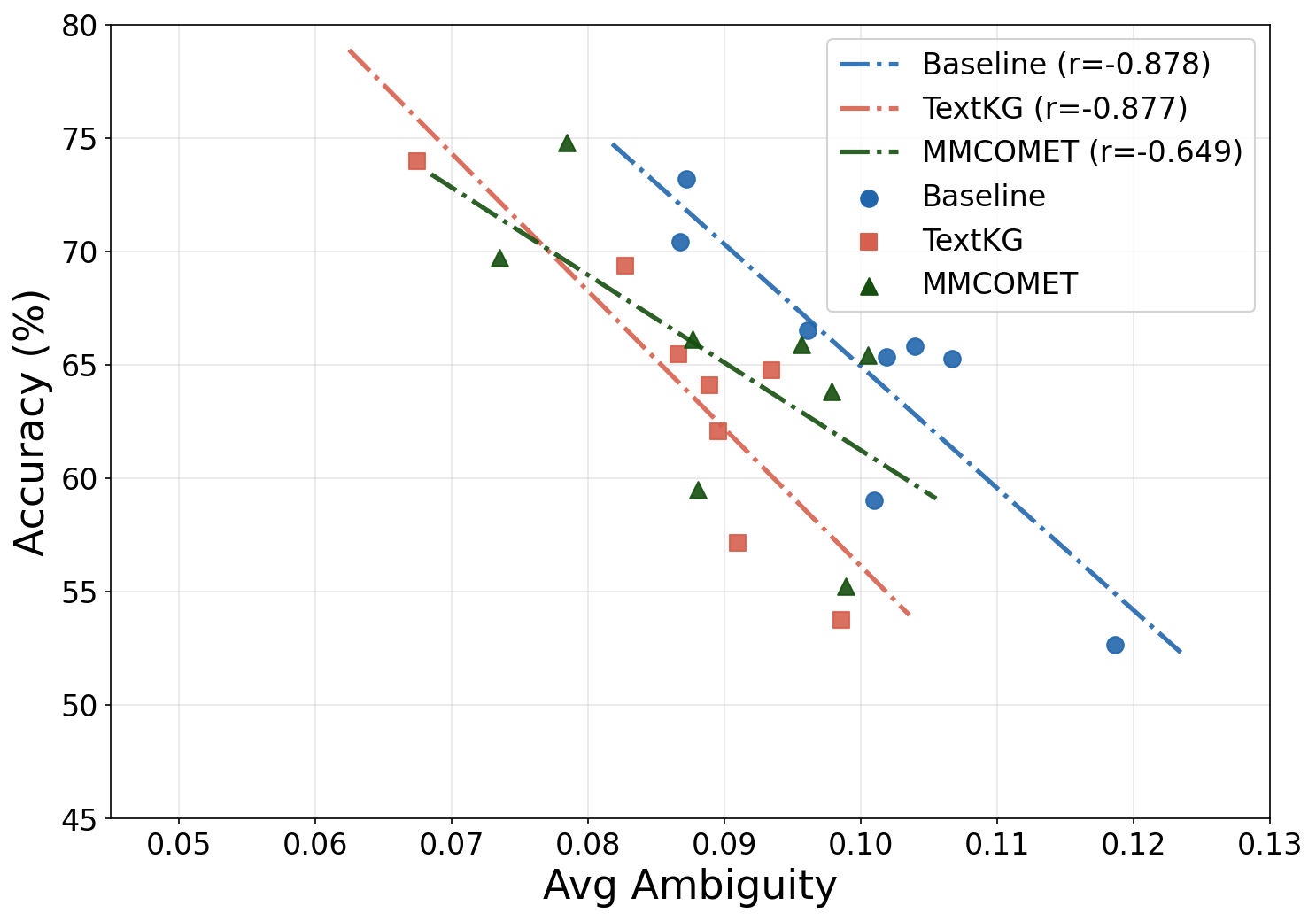}
\hfill
    \includegraphics[width=0.48\textwidth]{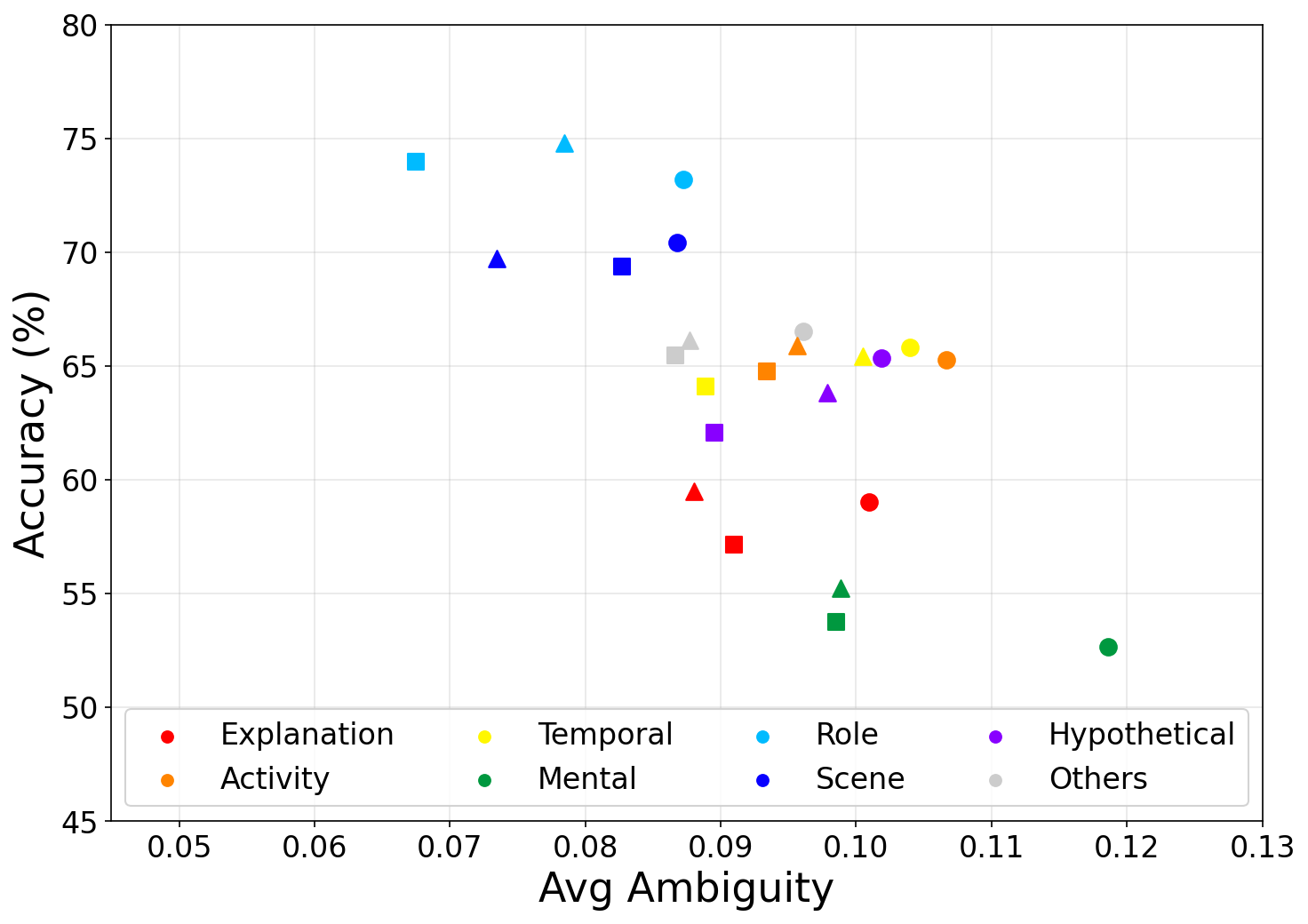}
\caption{Ambiguity vs Accuracy by Method (Left) and Category (Right).}
\label{fig: ambiguity_accuracy}
\end{figure*}

\begin{figure*}[!h]
\centering
\begin{minipage}[t]{0.33\textwidth}
\centering
\includegraphics[width=\textwidth]{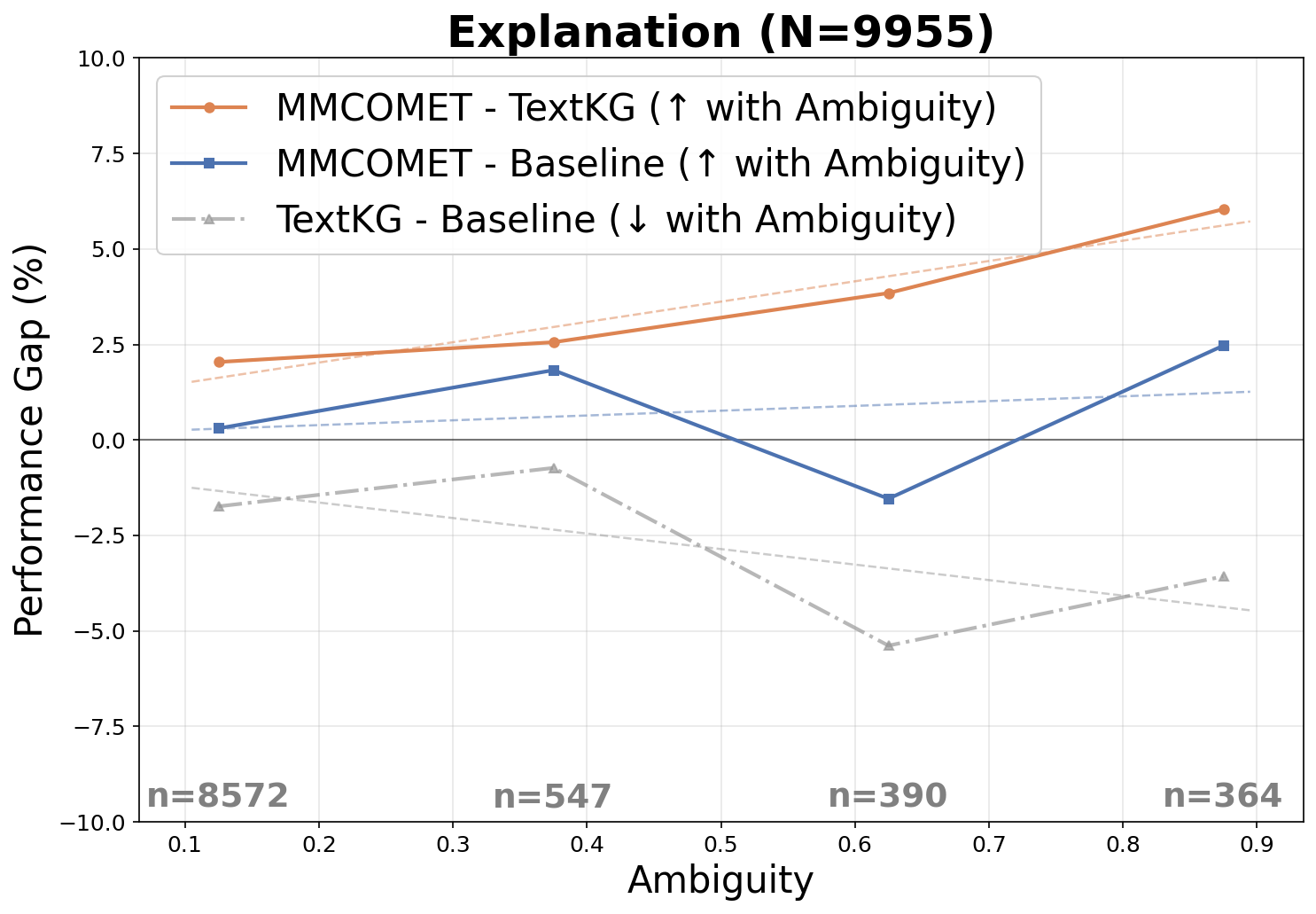}
\end{minipage}
\begin{minipage}[t]{0.32\textwidth}
\centering
\includegraphics[width=\textwidth]{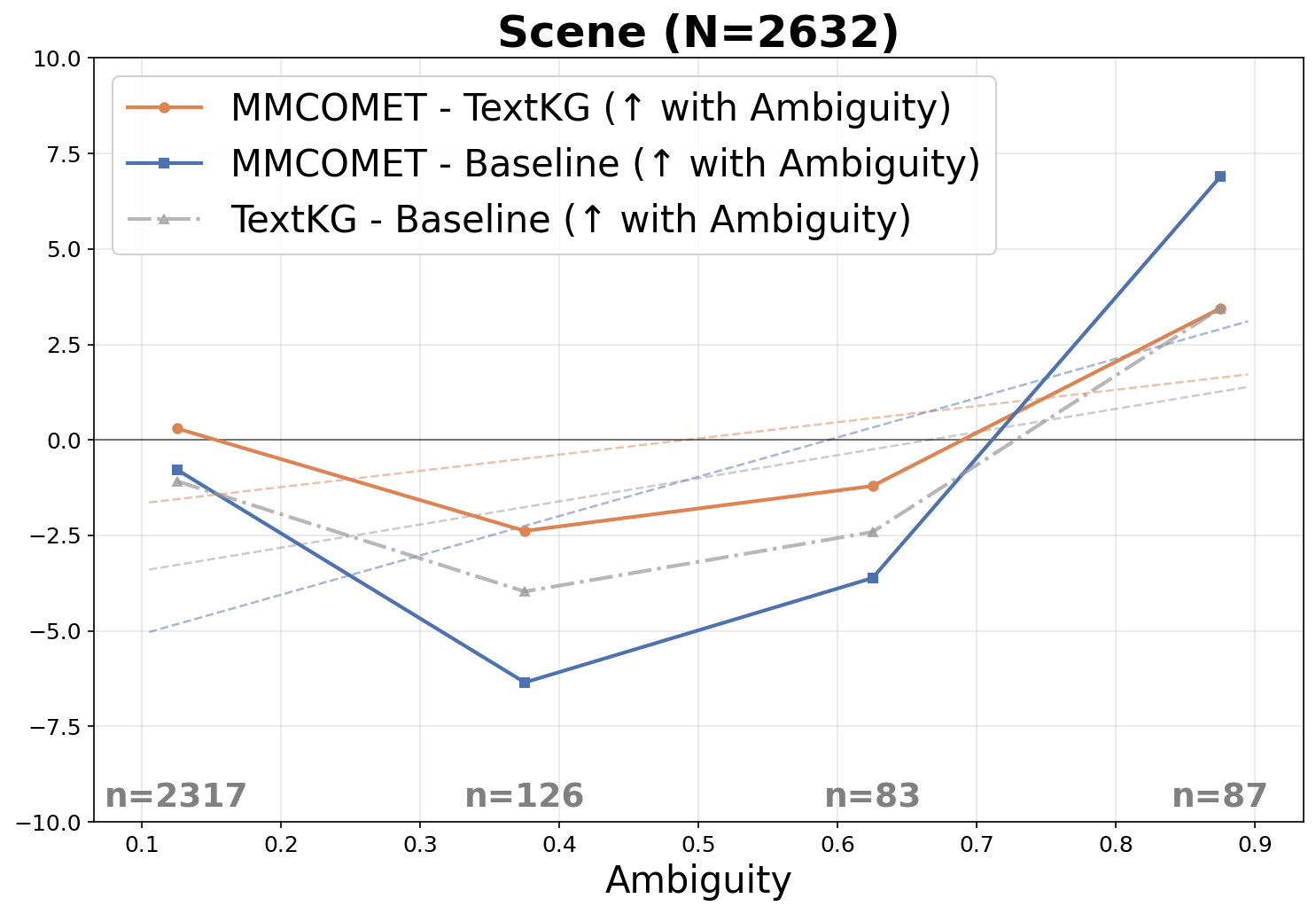}
\end{minipage}
\begin{minipage}[t]{0.32\textwidth}
\centering
\includegraphics[width=\textwidth]{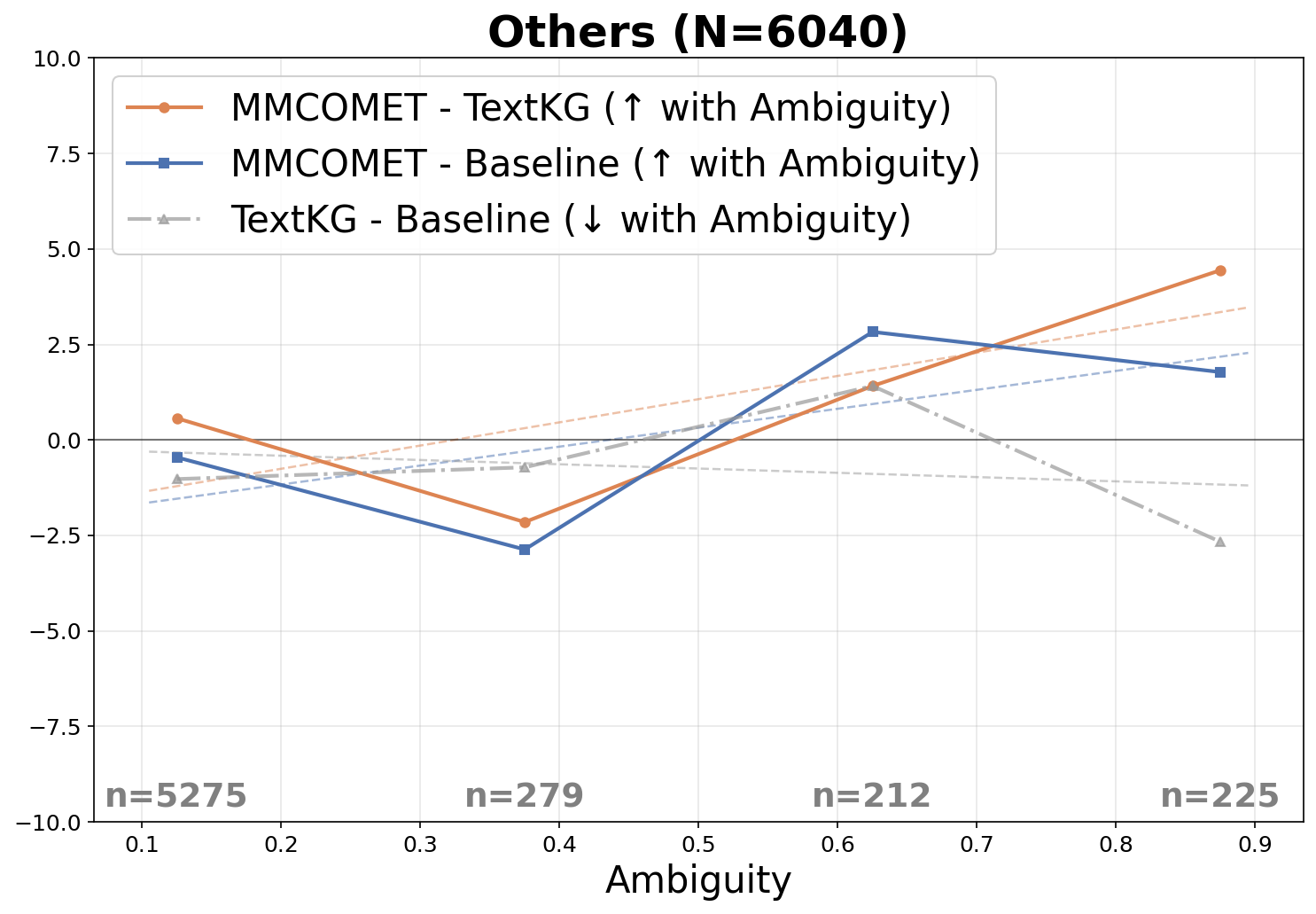}
\end{minipage}
\caption{Ambiguity vs performance gap for representative categories.}
\label{fig:performance_gap_category}
\end{figure*}
\paragraph{Ambiguity Analysis}
Figure~\ref{fig: ambiguity_accuracy} shows the correlation between average ambiguity and accuracy per category for each method. While Baseline and TextKG exhibit strong negative correlation ($r=-0.878$ and $r=-0.877$, respectively), MMCOMET shows a notably weaker correlation ($r=-0.649$), indicating that multimodal grounding partially mitigates the performance degradation associated with high ambiguity. Together, these results suggest that MMCOMET is particularly effective in challenging scenarios, providing more reliable commonsense augmentation precisely when the model is most uncertain. Also it reveals that across most categories including Explanation, Mental, and Activity, MMCOMET reduces average ambiguity while simultaneously improving accuracy compared to the baseline, suggesting that multimodal commonsense knowledge helps the model become more confident and more correct across diverse question types. In contrast, TextKG tends to reduce ambiguity but also degrades accuracy in many categories, indicating that TextKG often acts as noise.

Figure~\ref{fig:performance_gap} shows how the performance gap between methods varies across ambiguity bins. As ambiguity increases, the gap between MMCOMET and TextKG grows consistently, reaching about +4\% in the highest ambiguity bin, while the gap between MMCOMET and Baseline also increases monotonically. In contrast, the gap between TextKG and the baseline widens in the negative direction, indicating that TextKG becomes increasingly detrimental as the sample difficulty increases. This asymmetry suggests that visual grounding is especially critical for harder samples, where text-only signals fail to identify contextually relevant knowledge. We present the performance gap across representative categories in Figure~\ref{fig:performance_gap_category}. Across Explanation, Scene, and Others, the gap between MMCOMET and TextKG generally increases with ambiguity, consistent with the trend observed in the total analysis. While some intermediate bins show performance drops, we attribute this largely to the uneven sample distribution across bins rather than a systematic degradation.

\subsection{Domain Coverage}
\label{appendix:domain_coverage}
\begin{table}[!ht]
    \centering
    \footnotesize
    \caption{Domain contribution to retrieved phrases and leave-one-out performance impact, for VCR, VQA, and Image Captioning. P/S/E denote Physical, Social, Eventive.}
    \resizebox{\textwidth}{!}{%
    \begin{tabular}{c|ccc|cc|cc|ccc}
    \hline
    \multirow{3}{*}{\textbf{Domain}} & \multicolumn{3}{c|}{\textbf{Phrase Share (\%)}} & \multicolumn{7}{c}{\textbf{Leave-one-out Performance}} \\
    \cline{2-11}
    & \multirow{2}{*}{\textbf{VCR}} & \multirow{2}{*}{\textbf{VQA}} & \multirow{2}{*}{\textbf{IC}}
    & \multicolumn{2}{c|}{\textbf{VCR}} & \multicolumn{2}{c|}{\textbf{VQA}} & \multicolumn{3}{c}{\textbf{Image Captioning}} \\
    \cline{5-11}
    & & & & \textbf{InternVL38B} & \textbf{Qwen8B} & \textbf{Gemma31B} & \textbf{Qwen8B} & \textbf{BLEU-4} & \textbf{CIDEr} & \textbf{ROUGE-L} \\
    \hline
        P & 20.71 & 40.66 & 40.87 & \textbf{75.3} & \textbf{66.5} & 65.65          & 69.13          & 0.169          & 0.357          & 0.443 \\
        S & 56.13 & 46.96 & 53.56 & 74.9          & 66.4          & 66.05          & 69.68          & 0.173          & \textbf{0.372} & \textbf{0.448} \\
        E & 54.04 & 35.06 & 32.54 & 75.1          & 66.3          & 66.1           & 68.75          & \textbf{0.174} & 0.362          & 0.447 \\
    \hline
        \textit{All (Full)} & - & - & - & \textbf{75.3} & 66.4 & \textbf{66.31} & \textbf{69.74} & 0.17 & \textbf{0.372} & 0.446 \\
    \hline
    \end{tabular}%
    }
    \label{tab:domain_coverage}
\end{table}
We extend our analysis to quantify downstream gains based on joint domain coverage for VCR and VQA covering discriminative tasks, and for Image Captioning covering the generative task. The distribution of the retrieved phrases based on their associated domain/s is provided in Table~\ref{tab:domain_coverage}. We note that each phrase may belong to one or more domain, resulting in overlapping distributions.

For VCR, excluding the physical domain showed no performance loss. The domain distribution of the retrieved phrases for each task (Table \ref{tab:domain_coverage}) reveals that the VCR dataset heavily relies on the Social and Eventive domains, with more than 50\% of phrases. In comparison, the Physical domain accounts for only around 20\%. This demonstrates the significance of MMCOMET's social and eventive dimensions for visual reasoning. The VQA task and dataset show a more evenly distributed domain distribution, resulting in better performance when all domains are combined. For Image Captioning, excluding the physical domain degrades performance most, despite the social domain matching more than 50\% of phrases. This is because captions must ground in visible image components, regardless of depicted social aspects.

\subsection{Alignment Strategy}
\label{appendix:alignment_strategy}
\begin{wraptable}[8]{r}{0.72\textwidth}
\vspace{-12pt}
    \centering
    \footnotesize
    \caption{Alignment Strategy Ablation Study for VCR and Image Captioning downstream tasks.}
    \begin{tabular}{cc|c|ccc}
    \hline
        \multicolumn{2}{c|}{\textbf{Alignment}}  & \textbf{VCR} & \multicolumn{3}{c}{\textbf{Image Captioning}} \\
        \textbf{Similarity} & \textbf{Web Only} & \textbf{Accuracy} & \textbf{BLEU-4} & \textbf{CIDEr} & \textbf{ROUGE-L} \\ \hline
        o & x & 64.4 & 0.169 & 0.371 & 0.445 \\
        x & o & 63.1 & \textbf{0.177} & \textbf{0.387} & \textbf{0.452} \\
        o & o & \textbf{66.4} & 0.17 & 0.372 & 0.446 \\ \hline
    \end{tabular}%
    \label{tab:alignment_strategy}
\end{wraptable}
Table \ref{tab:alignment_strategy} isolates each alignment method's benefits using Qwen3-8B. For VCR, removing similarity-matched images (focused on visually concrete concepts) hurts performance most. Conversely, Image Captioning benefits more from abstract social or eventive concepts via web-based retrieval, improving without caption-sourced images. 
These results highlight the varying requirements of different downstream tasks, which benefit from distinct visual grounding routes. MMCOMET's hybrid construction pipeline provides broad coverage, encompassing both concrete and abstract concepts, rather than guaranteeing a task-specific optimum.

\subsection{Effect of Knowledge Source}
\label{appendix:llm_gen_figure}
\begin{table}[!h]
    \centering
    \footnotesize
    \vspace{-0.8em}
    \caption{Performance comparison against VTKG-C for VCR, VQA, Image Captioning(CIDEr) downstream tasks.}
    \begin{tabular}{c|cc|cc|cc}
    \hline
        \multirow{2}{*}{\textbf{Setup}} & \multicolumn{2}{c|}{\textbf{VCR}} & \multicolumn{2}{c|}{\textbf{VQA}}& \multicolumn{2}{c}{\textbf{IC}} \\
        & \textbf{InternVL38B} & \textbf{Qwen8B} & \textbf{Qwen8B} & \textbf{Gemma31B} & \textbf{Qwen8B}& \textbf{Qwen32B}\\ \hline
        Baseline & 73.8 & 65.3 & 68.60 & 64.72 & 0.232 & 0.151 \\ 
        TextKG & 74.1 & 62.7 & 68.41 & 65.55 & 0.346 & 0.147\\ 
        VTKG-C & 73.8 & 64.1 & \textbf{70.01} & 64.85 & 0.323 & 0.124 \\ 
        MMCOMET & \textbf{75.3} & \textbf{66.4} & 69.74 & \textbf{66.31} & \textbf{0.372} & \textbf{0.157}\\ \hline
    \end{tabular}%
    \label{tab:vtkg_vcr_vqa}
    \vspace{-1em}
\end{table}
\vspace{-5pt}
\begin{figure*}[!h]
    \centering
    \includegraphics[width=\linewidth]{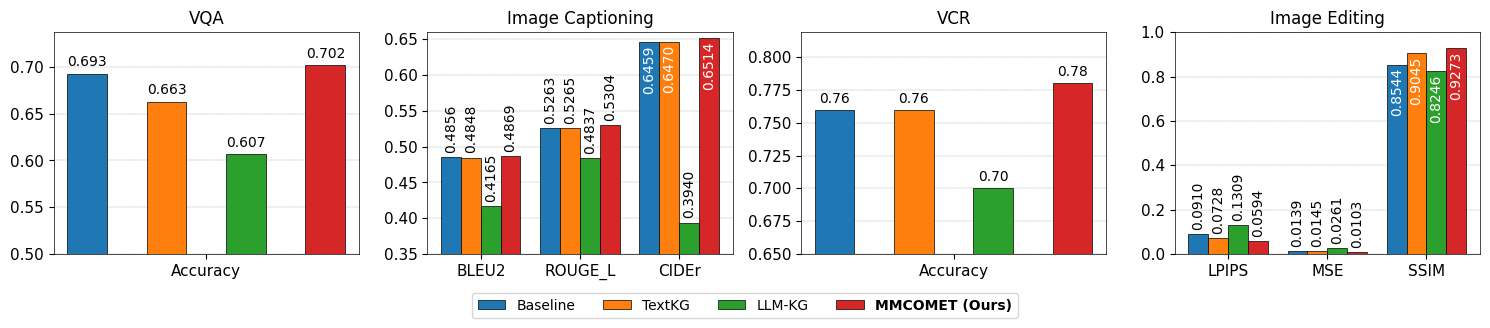}
    \caption{Incorporation of MMCOMET against baseline, TextKG, and LLM generated knowledge.}
    \label{fig:ablation_llm_gen}
\vspace{-15pt}

\end{figure*}
\paragraph{LLM-Generated Knowledge}As discussed in Sec.~\ref{sec:llm_ablation}, we compare MMCOMET against knowledge generated by prompting InternVL3-8B to produce commonsense phrases about objects, people, relationships, behaviors, and the contextual background of an input image, using 100 samples per task. Figure~\ref{fig:ablation_llm_gen} shows that MMCOMET outperforms all other setups for VQA and VCR, where model reasoning benefits from the integrated physical, social, and eventive knowledge. LLM-generated knowledge performs worst across all tasks, indicating that the structured knowledge from the commonsense graphs better directs visual grounding than naive knowledge integration. Perception-dominated tasks such as IC and IE benefit less from commonsense knowledge integration overall; however, MMCOMET still achieves greater gains than text-only knowledge, demonstrating more robust knowledge.

\paragraph{VTKG-C} We evaluate MMCOMET against VTKG-C, an existing multimodal commonsense knowledge graph limited to the physical domain, on three representative settings that jointly span the task space: VCR (reasoning-heavy), VQA (more general, perception-oriented), and Image Captioning (generative). Table~\ref{tab:vtkg_vcr_vqa} report results for VCR/VQA and Image Captioning, respectively. Injecting knowledge from VTKG-C does not benefit VCR, resulting in similar or even degraded performance compared to the baseline. On VQA, VTKG-C yields a marginal gain over MMCOMET for Qwen8B (70.01 vs. 69.74) but underperforms for Gemma31B, suggesting its physical-domain knowledge occasionally aligns with VQA's more perception-oriented questions but does not generalize across models. For Image Captioning, VTKG-C improves upon the baseline but falls short when compared to using the text-based TextKG method. Overall, MMCOMET outperforms VTKG-C on the majority of tasks and models, demonstrating the importance of extending commonsense knowledge beyond the physical domain to include social and eventive dimensions.

\section{Extrinsic Qualitative Analysis}
\label{appendix:qa_extrinsic}
To further illustrate the effect of using MMCOMET as auxiliary information in visual-language downstream tasks, we show a selection of representative examples in the following sections.

\begin{figure*}[!htbp]
    \centering
    \includegraphics[width=\textwidth]{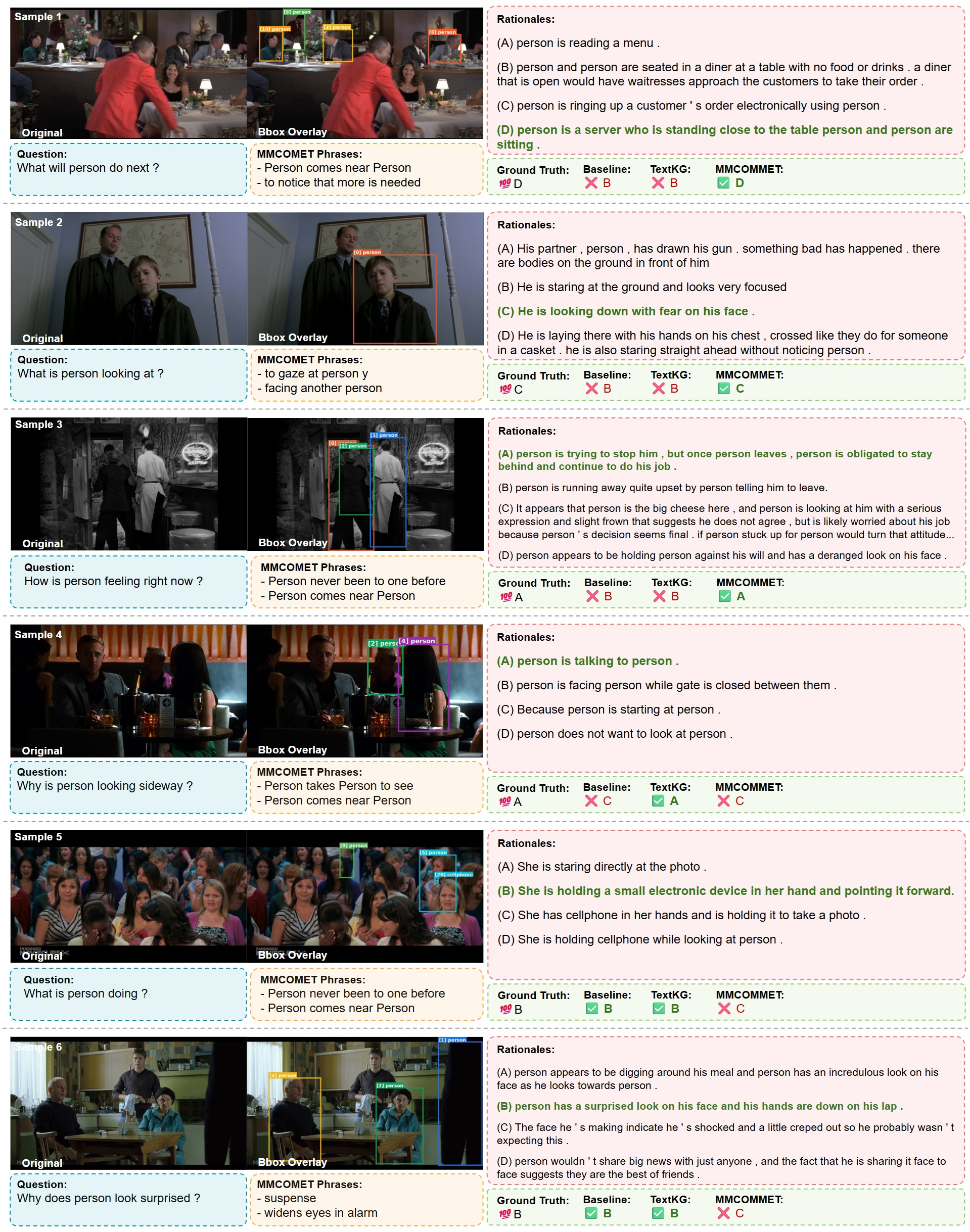}
    \caption{Qualitative samples for Visual Commonsense Reasoning.}
    \label{fig:full_qa_vcr}
\end{figure*}
\subsection{Visual Commonsense Reasoning}
\label{appendix:qa_vcr}

Figure~\ref{fig:full_qa_vcr} shows representative examples of VCR tasks, where the upper three cases are success cases and the lower three cases are failure cases. Taken together, these examples reveal that baseline models typically select rationales based on salient features. In contrast, MMCOMET output demonstrates richer commonsense grounding that can be grouped into two reasoning patterns. \textbf{(1) Perceptual state inference.} MMCOMET enables the model to reason about internal states that are not directly readable from superficial features. For instance, in sample 2, both the Baseline and TextKG fail to identify the correct rationale, selecting answers that reflect only surface-level scene description. MMCOMET, guided by retrieved phrases encoding perceptual attention of correctly infer based on the emotional quality underlying that gaze. \textbf{(2) Social and emotional context modeling.} Enhanced outputs incorporate implicit interpersonal dynamics that require reading beyond individual characters. In such cases, MMCOMET leverages phrases that capture proximity and relational roles to correctly identify the situation, discriminate among visually similar but socially distinct rationales, and arrive at the interpretation that best reflects the underlying situation. 

Similarly, MMCOMET's failures tend to fall into two patterns. First, retrieved phrases are broadly relevant to the image yet semantically misapplied, for example, in sample 6, the phrase `widens eyes in alarm' is successfully retrieved in a surprise scenario, but the model nonetheless selects an incorrect rationale centered on social inference, suggesting that retrieval relevance does not guarantee correct integration into downstream reasoning. Second, in cases where the question is based on fine-grained object-level or action-level details, models fail to connect meaningfully with the visual evidence in the image.  

\subsection{Visual Question Answering}
\label{appendix:qa_vqa}
\begin{figure*}[!htbp]
    \centering
    \includegraphics[width=\textwidth]{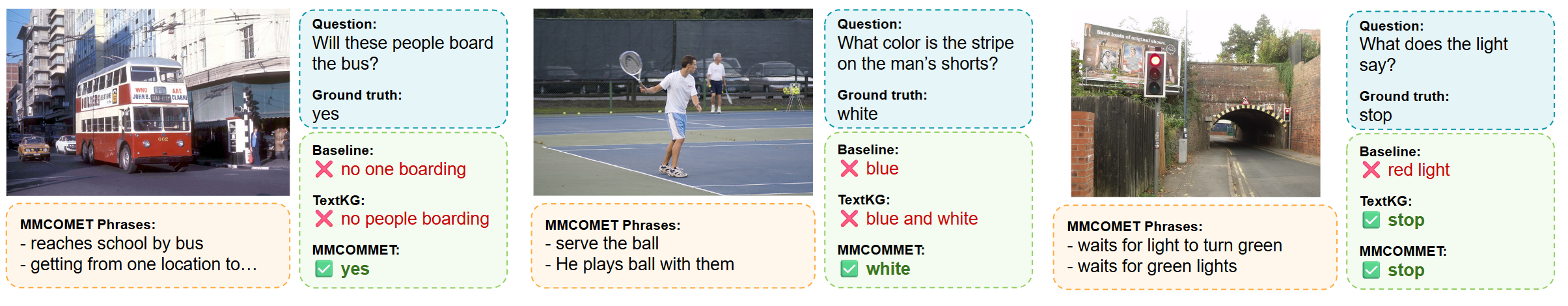}
    \caption{Qualitative samples for Visual Question Answering.}
    \label{fig:full_qa_vqa}
\end{figure*}
Figure~\ref{fig:full_qa_vqa} presents representative VQA examples where both Baseline and TextKG produce incorrect answers while MMCOMET succeeds. These cases illustrate how commonsense retrieval supports correct inference. 

In the middle example, the retrieved phrases`serve the ball' and `He plays ball with them' do not directly address the question about color. However, by anchoring the model to the correct activity context, these phrases appear to facilitate more focused visual attention, enabling the model to localise the relevant region and correctly identify its color. This suggests that commonsense retrieval can also benefit the model by providing contextual scaffolding that sharpens the focus on question-relevant visual details.

In the right example, it highlights a complementary failure mode of the baseline. When asked what the traffic light says, the baseline responds with the visual feature only. However, both TextKG and MMCOMET correctly answer stop, suggesting access to commonsense knowledge about traffic light conventions. It seems that the retrieved commonsense phrases `wait for green lights' and `waits for light to turn green' help the model interpret the question at the appropriate level of abstraction, shifting from perceptual description to functional meaning.

\begin{figure*}[!t]
    \centering
    \includegraphics[width=0.8\linewidth]{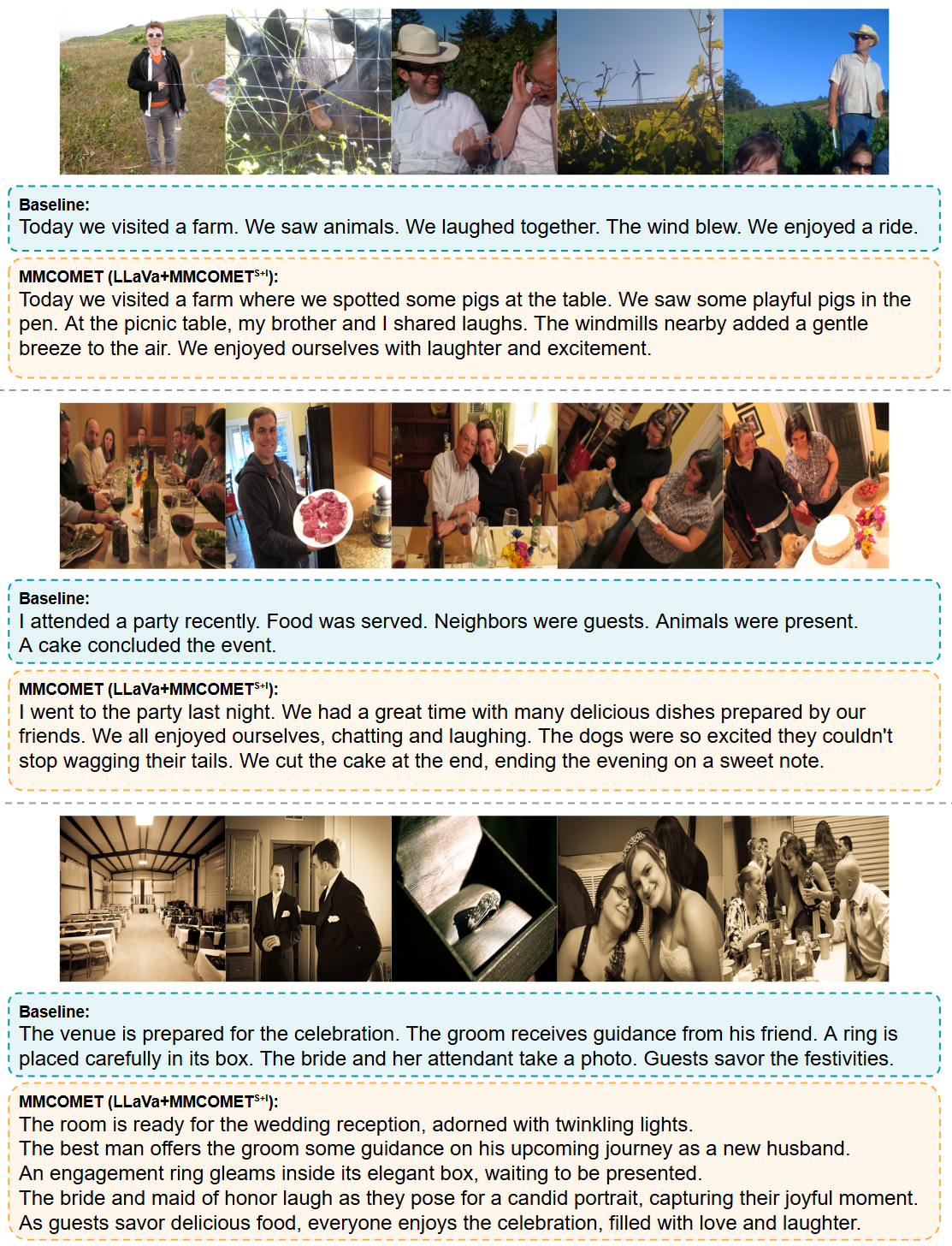}
    \caption{Qualitative samples for Visual Storytelling.}
    \label{fig:full_qa_vst}
\end{figure*}
\subsection{Visual Story Telling}
\label{appendix:qa_vst}
Figure~\ref{fig:full_qa_vst} presents representative Visual storytelling examples. Baseline outputs tend to enumerate observed visual elements as isolated facts, producing narratives that are factually grounded but experientially flat. MMCOMET, in contrast, introduce richer narrative structure by embedding social and eventive context around the same visual observations.

The distinction is evident across all three examples. In the farm visit story, the baseline notes the presence of animals and a ride. At the same time, MMCOMET reconstructs the scene as a shared experience, situating individual observations within a coherent narrative of the outing. In the party example, the baseline lists food, guests, and a concluding cake. In contrast, MMCOMET elaborates on the social atmosphere, capturing collective enjoyment through details such as chatting, laughing, and excited dogs. Most notably, in the wedding story, MMCOMET transforms each visual panel into a narratively connected moment that is not directly observable from the images alone.

Taken together, these examples suggest that commonsense retrieval enables the model to move beyond visual enumeration toward socially and eventively grounded storytelling, producing narratives that better reflect how people experience and recall shared events.

\subsection{Image Editing}
\label{appendix:qa_ie}
\begin{figure*}[!htbp]
    \centering
    \includegraphics[width=\linewidth]{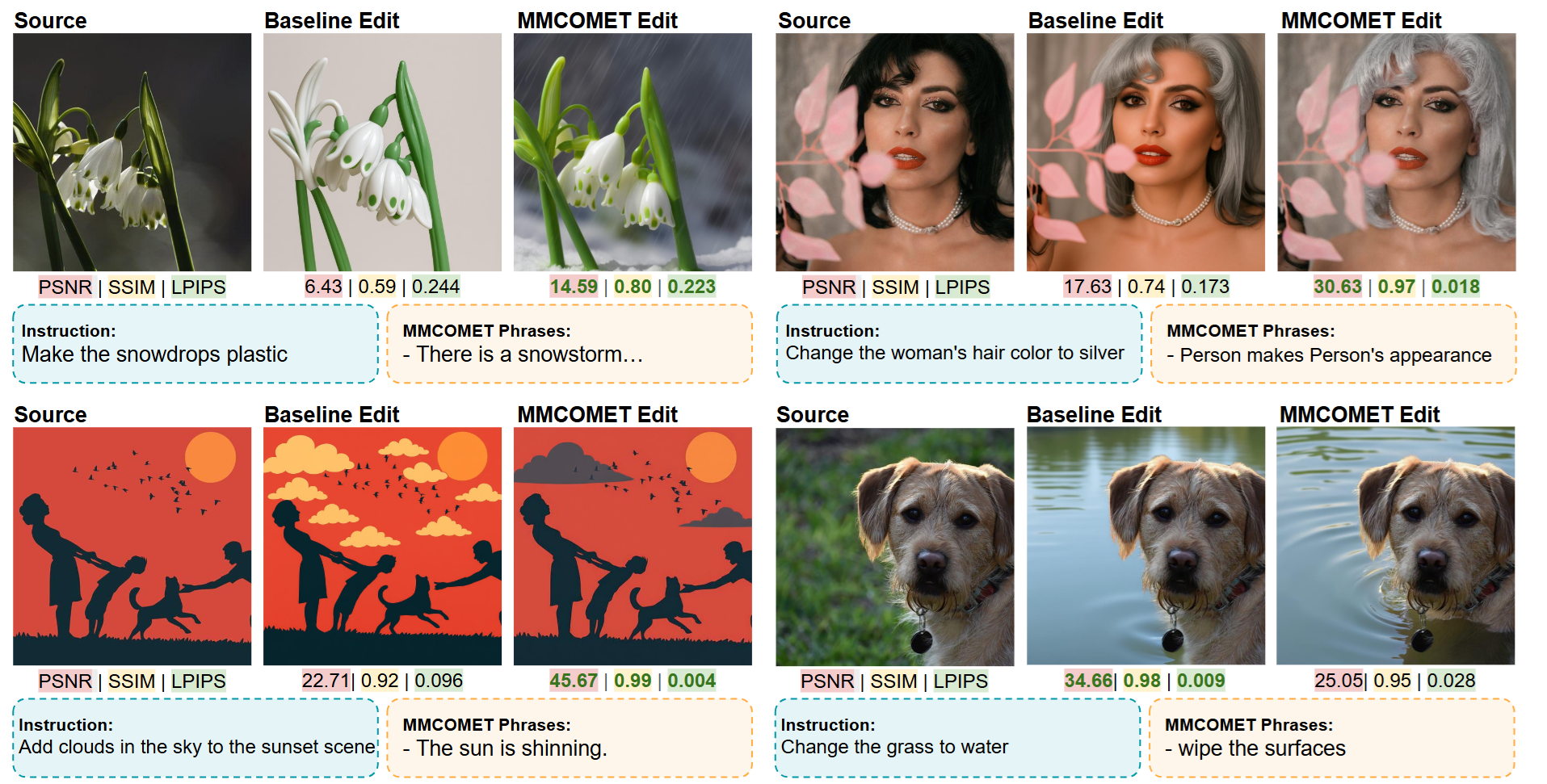}
    \caption{Qualitative samples for Image Editing.}
    \label{fig:full_qa_img_edit}
\end{figure*}
Figure~\ref{fig:full_qa_img_edit} presents representative image editing examples. Across all cases, MMCOMET consistently achieves better source preservation than the baseline, as reflected in higher PSNR and SSIM scores, while applying targeted modifications that align with the instruction's intent (e.g., correctly localising appearance changes to the relevant region while leaving surrounding content intact in the upper-right sample). This pattern suggests that commonsense retrieval encourages conservative editing, prioritising structural fidelity over aggressive transformation. In the case of `Change the grass to water', the retrieved phrase `wipe the surfaces' provides only a generic surface-level association, failing to capture the material and textural transformation required. Consequently, MMCOMET falls short of edit fidelity in the automatic metric, yet still introduces subtle surface undulations that are more consistent with the appearance of water than the baseline output, suggesting that, even in failure cases, the model attends to semantically relevant visual properties rather than defaulting to arbitrary modifications.

\subsection{Image Captioning}
\label{appendix:qa_ic}
\begin{figure*}[!htbp]
    \centering
    \includegraphics[width=\linewidth]{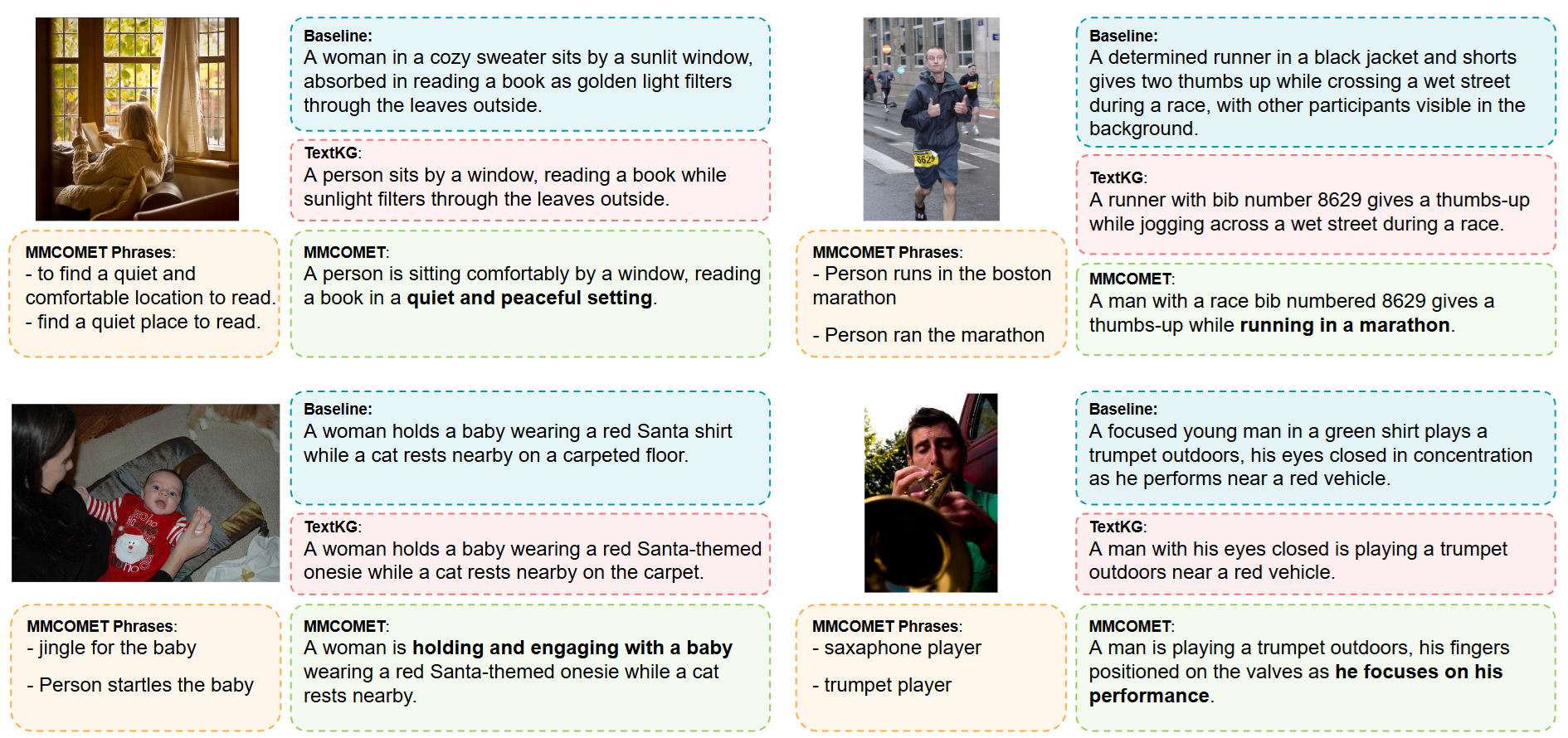}
    \caption{Qualitative samples for Image Captioning.}
    \label{fig:full_qa_ic}
\end{figure*}

Figure \ref{fig:full_qa_ic} presents representative image captioning examples. Baseline-generated captions tend to over-describe objects that are not the main focus of the image, such as "black jacket and shorts" in the top-right image and "near a red vehicle" in the bottom-right image. In contrast, MMCOMET's captions consistently move beyond surface-level visual description toward more contextually grounded characterisations. Rather than enumerating appearance and spatial arrangement, MMCOMET outputs capture the underlying intent, role, or interactional quality of the depicted scene. 

This pattern is evident in two complementary ways. In the marathon example, the baseline describes the runner's appearance and physical context in detail, yet fails to identify the event itself. MMCOMET, guided by retrieved phrases, correctly situates the action within a specific event schema, producing a caption that reflects both appearance and context. In the baby example, both the baseline and TextKG describe the scene as a static holding relationship, whereas MMCOMET reframes the interaction as holding and engaging with a baby, informed by retrieved phrases that encode interpersonal engagement with an infant.

\section{Intrinsic Qualitative Analysis}
\label{appendix:qa_intrinsic}
\subsection{Alignment Quality Analysis}
Table~\ref{tab: human_eval} shows the result of human evaluation on image-phrase pairs, evaluating alignment quality. 

\paragraph{Result by retrieval source }Web-based retrieval (WSE) consistently outperforms similarity matching over open-source caption datasets (OID). Aggregated across all relations, WSE achieves 91.5\% (FM+PM) and 79.0\% (FM), compared to 82.4\% and 61.2\% for OID. This difference reflects the advantage of query-sensitive search for abstract and compositional phrases, while caption datasets often lack explicit coverage of eventive or social expressions. Approximately 72\% of phrases in MMCOMET are grounded via web retrieval, underscoring the importance of the hybrid routing strategy.

\paragraph{Result by relation types}
Alignment quality also varies across relation types. 
Physical relations achieve the highest grounding rates (97.8\% FM+PM), followed by eventive (83.3\%) and social (84.3\%) relations. This trend aligns with phrase concreteness: tangible entities are easier to visually ground, whereas abstract emotional or intentional states exhibit greater ambiguity. Despite this inherent difficulty, even the most challenging relation types maintain alignment above 77\% (FM+PM). 
\subsection{Fully vs. Partially Matched Images} \label{appendix:full_vs_partial}

\begin{figure*}[!t]
  \centering  \includegraphics[width=0.8\linewidth]{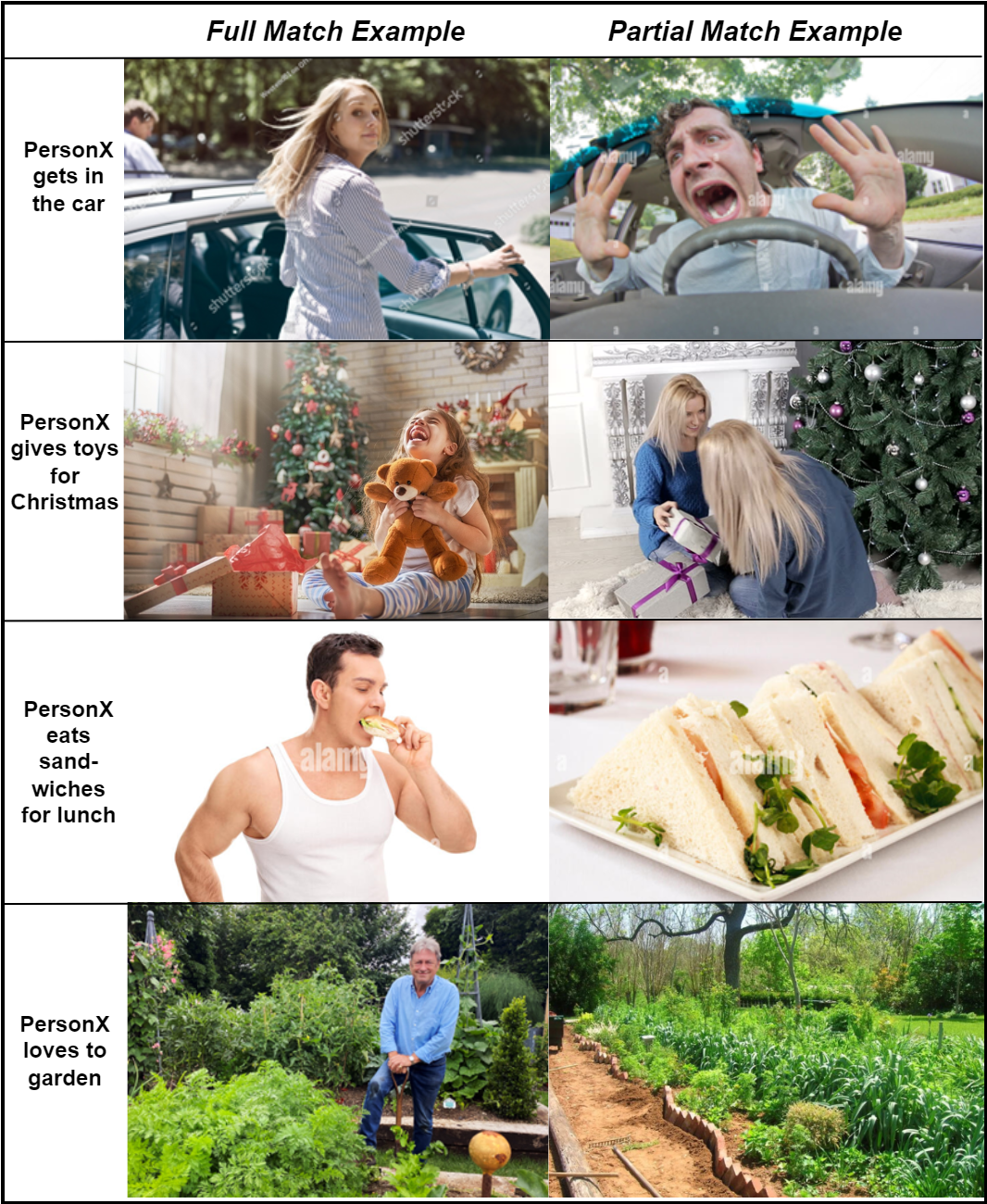}
  \caption{Examples of a fully matched image (left) and partially matched image (right) for each 4 different commonsense phrases.}
  \label{fig: full_vs_partial}
\end{figure*}

Figure \ref{fig: full_vs_partial} shows examples of what we consider a full versus partial match for an image. An image is a full match if all the actions and concepts in the commonsense phrase are presented in the image. On the other hand, it is a partial match if some aspect is missing but the overall theme of the image still matches the commonsense phrase and will therefore still be useful if applied to the downstream task. In the first (`\textit{PersonX gets in the car}') and third example (`\textit{PersonX eats sandwiches for lunch}'), the left image is considered a full match as the images clearly show the action of a person entering a car and eating a sandwich respectively. Conversely, the images on the right are considered partial matches as for the first example, the  image on the right just shows a person in the car but the action of `getting into the car' is missing. Likewise, for the third example, the image on the right simply depicts sandwiches but the action of `eating' is absent. For the second (`\textit{PersonX gives toys for Christmas'}) and fourth example (`\textit{PersonX loves to garden}'), an important noun or concept is missing from the partially matched images. Specifically, for the second example, although the action of `giving' is present in the image on the right, the tangible noun `toys' is not visually captured whereas the left image of a child holding a toy bear heavily implies she had received it from another person. Similarly, the image on the right in the fourth example is missing the important entity `person' whereas the left image is a full match as it clearly depicts a person happy in his garden. 

\subsection{Head-Relation-Tail Image Phrase Matching}
\label{appendix:relation_types}

\begin{figure*}
  \centering
  \includegraphics[height=0.35\textheight, keepaspectratio]{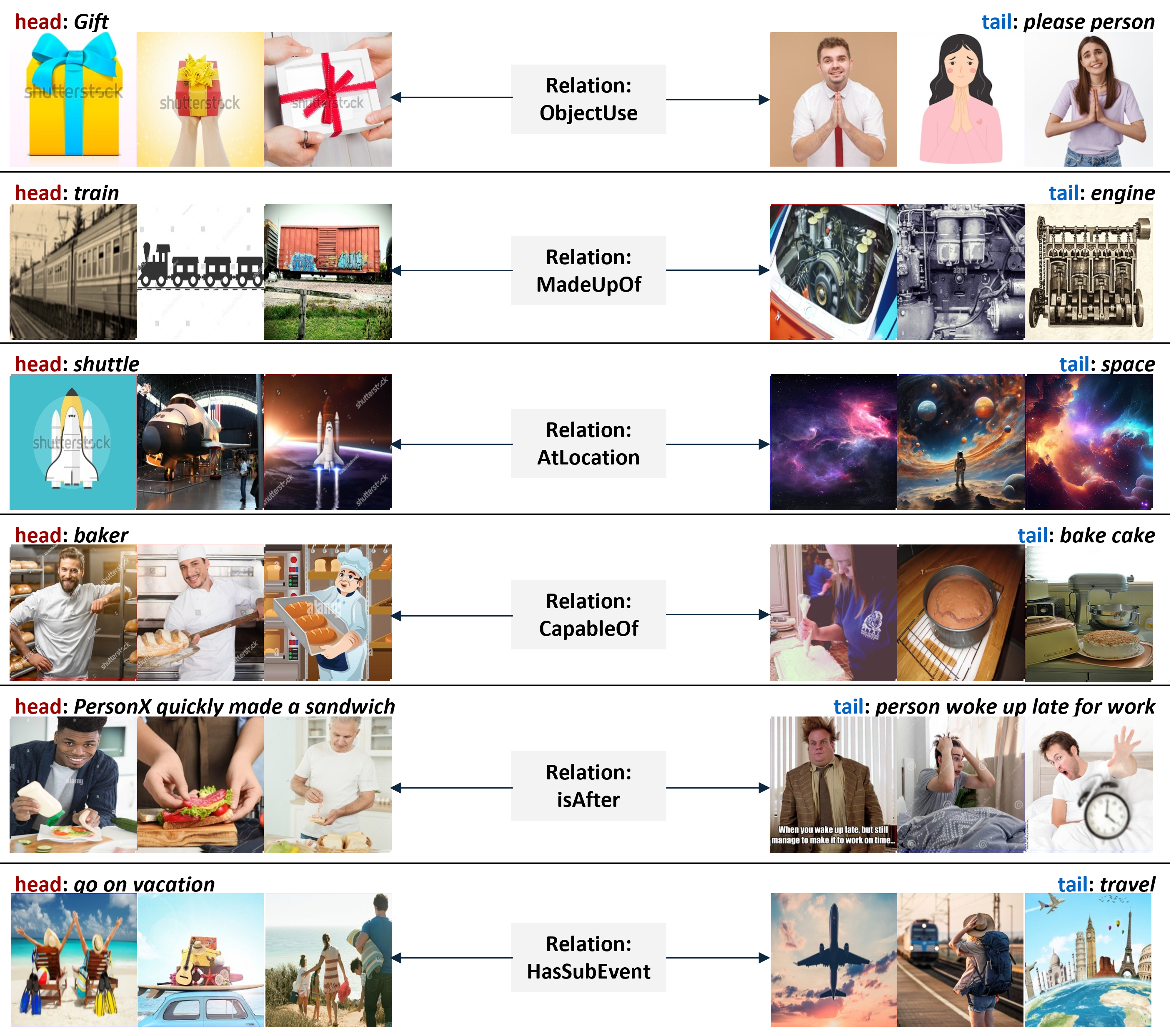}

  \vspace{0.3em}
  \includegraphics[height=0.35\textheight, keepaspectratio]{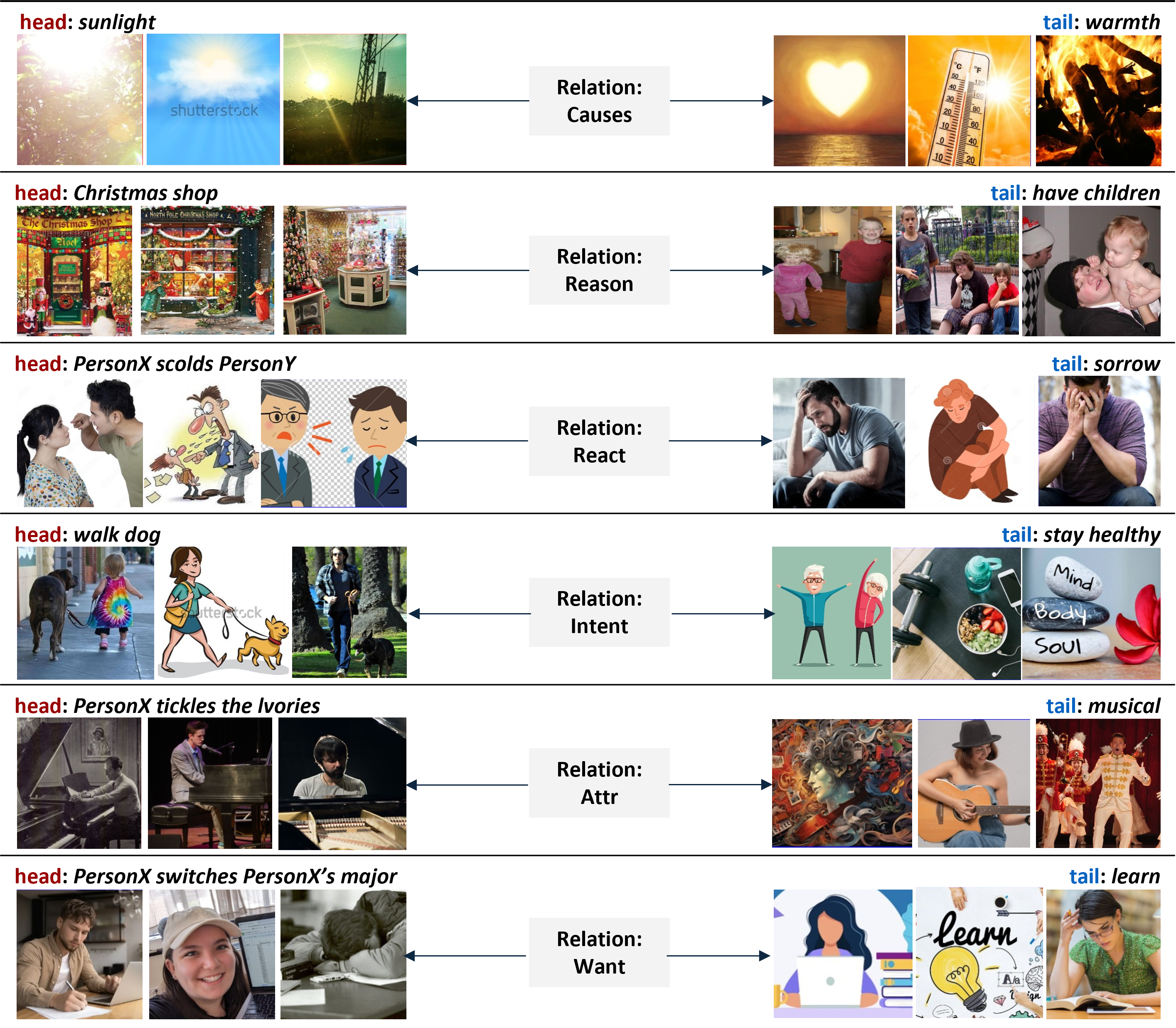}

  \caption{Examples of retrieved images corresponding to selected head and tail phrases across multiple relation types. Blue, orange and green arrows indicate physical, eventive and social relations, respectively. Images with red borders indicate images obtained from open-source datasets, while images with blue borders indicate images retrieved through web searches.}
  \label{fig: good_examples}
\end{figure*}

Figure~\ref{fig: good_examples} shows examples of several relations, each with 3 matching image examples for each of the head and tail are shown. For instance, connecting “\textit{gift}” and “\textit{please person}” through images depicting both the physical form of gifts and the emotional responses they elicit. Eventive relations such as \textit{isAfter} and \textit{HasSubEvent} demonstrate temporal and causal consistency, where sequences like “\textit{PersonX quickly made a sandwich}” and “\textit{PersonX woke up later for work}” visually manifest cause-and-effect coherence.

\newpage
\clearpage

\section{Adaptive Knowledge Utility Router}
\label{app:router}

To further address the heterogeneous reasoning requirements of diverse downstream vision-language tasks, we propose an adaptive knowledge utility router that evaluates the usefulness of different knowledge sources or routes. Instead of relying on hard, one-hot classification, which treats route utility as a binary switch, discarding complementary information from overlapping paths, the adaptive knowledge router models the utility of each route independently using a soft utility score. This multi-route formulation preserves overlapping successful routes, accounts for potentially harmful interventions, and allows the model to default to a no-knowledge prediction when additional knowledge yields no benefit. The adaptive knowledge routing framework has three stages (Algorithm \ref{alg:acur_train} and \ref{alg:acur_inference}): 1)~oracle-derived utility construction, 2) utility-router optimization, and 3) routed inference. 

\subsection{Router Component Construction}
During utility component construction, a frozen VLM model extracts all baseline and knowledge-augmented components for each sample. To exhaust the commonsense knowledge coverage of MMCOMET, we utilitze its individual domains and their combinations as 7 distinct routes $r \in \mathcal{R}_{K}$: Physical Only (P), Social Only (S), Eventive Only (E), P+S, P+E, S+E, and All. The baseline components includes the base prediction $\hat{y}_i^0$, the relevant token logits $p_i^0$, and the base confidence profile $s_i$, consisting of entropy, maximum predicted probability, and confidence margin. For each sample and knowledge route, we extract the augmented prediction $\hat{y}_i^r$, soft utility score $u_i^r$, and encoded state $z_i^r$. The soft utility score represents the net gain in the log-probability gap between the ground truth and the strongest incorrect answer across the base and the knowledge-augmented models. An independent utility encoder using a  Qwen3-
VL-Embedding-8B backbone constructs the encoded state that serves as a a route-specific representation derived from the input image, question, retrieved MMCOMET phrases, base confidence profile, and route information. The soft utility score and the encoded state serve as the training inputs and targets for training the knowledge utility router.

\subsection{Router Optimization and Inference}
\begin{wrapfigure}[18]{r}{0.35\textwidth}
    \vspace{-30pt}
    \centering
    \captionof{table}{Relevant Router Hyperparameters}
    \vspace{-10pt}
    \label{tab:router_parameters}
    \begin{tabular}{@{}lrl@{}}
    \toprule
    \multicolumn{1}{c}{\textbf{Hyperparameter}} & \multicolumn{1}{c}{\textbf{Value}} &  \\ \midrule
    \multicolumn{2}{c}{\textbf{Utility Encoder}} &  \\ \midrule
    hidden\_dim & 512 &  \\
    output\_dim & 512 &  \\ \midrule
    \multicolumn{2}{c}{\textbf{Adaptive Router}} &  \\ \midrule
    n\_head & 8 &  \\
    num\_layers & 3 &  \\
    dim\_feedforward & 1024 &  \\
    dropout & 0.1 &  \\
    learning\_rate & 1e-05 &  \\
    epochs & 100 &  \\
    early\_stop & 5 &  \\ \bottomrule
    \end{tabular}
\end{wrapfigure}
The knowledge utility router is a lightweight, Transformer-based model that jointly predicts the utility scores for all routes $u_i^{\mathcal{R}_K}$ for each sample within a minibatch. To mitigate the impact of outliers and extreme margin values during training, we constrain all utility scores using a hyperbolic tangent function parameterized by a temperature hyperparameter $\mathcal{T} \in \{2, 5, 10\}$ during training. Optimization is guided by a joint objective consisting of Huber loss for regression and a within-sample pairwise margin-ranking loss to effectively identify optimal routes. During inference, when the maximum predicted utility score falls below a threshold $\tau$ $\in \{-0.1, -0.01, 0, 0.01, 0.1\}$, the router bypasses external knowledge and returns the base answer, indicating that no knowledge is beneficial. Otherwise, knowledge from the highest-utility route is used to generate the final prediction. 

\begin{algorithm}[t]
\caption{Adaptive Knowledge Utility Router Training}
\label{alg:acur_train}
\small
\begin{algorithmic}[1]
\Require Training set $\mathcal{D}$, frozen VLM $M$, MMCOMET
$\mathcal{G}$, knowledge routes $\mathcal{R}_{K}$,
utility router $U_{\theta}$, scaling temperature $T$
\Ensure Trained utility router $U_{\theta}$

\Statex \textbf{Oracle-derived utility construction}

\ForAll{$(I_i,Q_i,y_i)\in\mathcal{D}$}

    \State $(\hat{y}_i^0,p_i^0)
    \gets M(I_i,Q_i)$

    \State $s_i
    \gets
    [H(p_i^0),\max(p_i^0), p_{i,(1)}^0-p_{i,(2)}^0]$

    \ForAll{$r\in\mathcal{R}_{K}$}

        \State $K_i^r
        \gets
        \textsc{Retrieve}
        (\mathcal{G},I_i,Q_i,r,B)$

        \State $(\hat{y}_i^r, p_i^r)
        \gets
        M(I_i,Q_i,K_i^r)$

        \State $u_i^r
        \gets
        \left[ \log p_i^r(y_i) - \max_{y'_i \neq y_i} \log p_i^r(y'_i) \right] - \left[ \log p_i^0(y_i) - \max_{y'_i \neq y_i} \log p_i^0(y'_i) \right]$

        \State $z_i^r
        \gets
        \textsc{Encode}
        (I_i,Q_i,K_i^r,s_i,r)$

    \EndFor
\EndFor

\Statex \textbf{Utility-router optimization}

\Repeat
    \ForAll{$(z_i^{\mathcal{R}_K},u_i^{\mathcal{R}_K})$ in a minibatch}
        \State $\hat{u}_i^{\mathcal{R}_K}
        \gets
        U_{\theta}(z_i^{\mathcal{R}_K})$
    \EndFor

    \State ${u'}_i^{\mathcal{R}_K}
    \gets \textsc{tanh}(u_i^{\mathcal{R}_K} / T)$

    \State $\mathcal{L}_{\mathrm{huber}}
    \gets \textsc{HuberLoss}({u'}_i^{\mathcal{R}_K}, \hat{u}_i^{\mathcal{R}_K})$

    \State $\mathcal{L}_{\mathrm{rank}}
    \gets \textsc{MarginRankingLoss}({u'}_i^{\mathcal{R}_K}, \hat{u}_i^{\mathcal{R}_K})$

    \State Update $\theta$ using
    $\mathcal{L}_{\mathrm{huber}}
    +\lambda_{\mathrm{rank}}\mathcal{L}_{\mathrm{rank}}$

\Until{convergence}

\State \Return $U_{\theta}$

\end{algorithmic}
\end{algorithm}

\begin{algorithm}[t]
\caption{Adaptive Knowledge Utility Router Inference}
\label{alg:acur_inference}
\small
\begin{algorithmic}[1]
\Require Input $(I,Q)$, frozen VLM $M$, MMCOMET $\mathcal{G}$,
router $U_{\theta}$, knowledge routes $\mathcal{R}_{K}$, utility threshold $\tau$
\Ensure Prediction $\hat{y}$

\State $(\hat{y}^{0},p^{0})
\gets M(I,Q)$

\State $s
\gets
[H(p^{0}),\max(p^{0}),p_{(1)}^{0}-p_{(2)}^{0}]$

\ForAll{$r\in\mathcal{R}_{K}$}

    \State $K^{r}
    \gets
    \textsc{Retrieve}(\mathcal{G},I,Q,r,B)$
    
    \State $z^{r}
    \gets
    \textsc{Encode}(I,Q,K^{r},s,r)$
    
\EndFor

\State $\hat{u}^{\mathcal{R}_K}
\gets
U_{\theta}(z^{\mathcal{R}_K})$

\State $r^{*}
\gets
\arg\max_{r\in\mathcal{R}_{K}}
\hat{u}^{r}$

\If{$\hat{u}^{r^{*}}\leq\tau$}
    \State \Return $\hat{y}^{0}$
    \Comment{No knowledge required}
\Else
    \State $\hat{y}
    \gets
    M(I,Q,K^{r^{*}})$
    \State \Return $\hat{y}$
\EndIf

\end{algorithmic}
\end{algorithm}

\subsection{Adaptive Routing Setup}

We evaluate our proposed adaptive router through a 5-fold cross-validation on MMStar\footnote{https://mmstar-benchmark.github.io/} (N=500) to ensure robustness. MMStar is a comprehensive VQA dataset that features six distinct question types with different knowledge requirements, including Coarse Perception (CP), Fine-grained Perception (FP), Instance Reasoning (IR), Logical Reasoning (LR), Math (MA), and Science and Technology (ST). We use different model families: Qwen3-VL-8B-Instruct, Qwen3-VL-32B-Instruct, LLaVa1.6-34B, and Mistral3.1-24B. For MMCOMET knowledge retrieval during each knowledge-augmented route, we maintain the same hyperparameters as outlined in Appendix \ref{appendix:eval_setup}. For the utility encoder, we use Qwen3-VL-Embedding-8B as the backbone to extract encodings for the image, question, knowledge phrases, route information, and confidence profile. Each component is individually encoded, then concatenated, and fused via linear layers. Table \ref{tab:router_parameters} outlines the relevant hyperparameters. We use accuracy as our primary performance metric.

\subsection{Adaptive Routing Results}

\begin{table}[t]
\centering
\scriptsize
\caption{MMStar categorical performance comparison between baseline, full MMCOMET, and MMCOMET with the adaptive router. CP: Coarse Perception; FP: Fine-grained Perception; IR: Instance Reasoning; LR: Logical Reasoning; MA: Math; ST: Science and Technology.}
\label{tab:mmstar_breakdown}
\begin{tabular}{@{}llrrrrrrr@{}}
\toprule
\textbf{Model} & \textbf{Setup} & \multicolumn{1}{l}{\textbf{Overall}} & \multicolumn{1}{c}{\textbf{CP}} & \multicolumn{1}{c}{\textbf{FP}} & \multicolumn{1}{c}{\textbf{IR}} & \multicolumn{1}{c}{\textbf{LR}} & \multicolumn{1}{c}{\textbf{MA}} & \multicolumn{1}{c}{\textbf{ST}} \\ \midrule
Qwen3-32B & Baseline & 0.7160 & 0.7614 & \textbf{0.7412} & \textbf{0.8161} & 0.6912 & 0.7500 & \textbf{0.5341} \\
 & MMCOMET (Full) & 0.7000 & 0.7500 & 0.7176 & 0.7931 & 0.7206 & 0.7024 & 0.5227 \\
 & MMCOMET + Router & \textbf{0.7260} & \textbf{0.7841} & \textbf{0.7412} & 0.8046 & \textbf{0.7353} & \textbf{0.7619} & \textbf{0.5341} \\ \midrule
LLaVa1.6-34B & Baseline & 0.5060 & \textbf{0.6364} & \textbf{0.4588} & 0.6552 & 0.4265 & \textbf{0.4881} & \textbf{0.3523} \\
 & MMCOMET (Full) & 0.4720 & 0.6023 & 0.3882 & 0.6437 & 0.4706 & 0.4405 & 0.2841 \\
 & MMCOMET + Router & \textbf{0.5080} & \textbf{0.6364} & 0.4000 & \textbf{0.7011} & \textbf{0.5000} & 0.4762 & 0.3295 \\ \midrule
Mistral & Baseline & \textbf{0.6020} & 0.6818 & 0.5176 & \textbf{0.7241} & 0.5735 & 0.6071 & \textbf{0.5000} \\
 & MMCOMET (Full) & 0.5840 & 0.6932 & 0.5294 & \textbf{0.7241} & 0.5441 & 0.5952 & 0.4091 \\
 & MMCOMET + Router & \textbf{0.6020} & \textbf{0.7045} & \textbf{0.5412} & 0.6897 & \textbf{0.5882} & \textbf{0.6310} & 0.4545 \\ \midrule
Qwen3-8B & Baseline & 0.6460 & \textbf{0.7386} & \textbf{0.6824} & 0.6897 & 0.6176 & 0.6548 & 0.4886 \\
 & MMCOMET (Full) & 0.6300 & 0.7045 & 0.6118 & \textbf{0.7471} & 0.6029 & 0.6548 & 0.4545 \\
 & MMCOMET + Router & \textbf{0.6520} & 0.7159 & 0.6706 & 0.6897 & \textbf{0.6324} & \textbf{0.6905} & \textbf{0.5114} \\ \bottomrule
\end{tabular}
\end{table}

Table \ref{tab:mmstar_breakdown} shows the overall and per-category accuracy between the no-knowledge baseline, full MMCOMET integration, and MMCOMET with the proposed adaptive, dynamic router.  On this dataset, the full MMCOMET framework degrades overall baseline performance, primarily driven by the heterogeneity of question types. Despite this, MMCOMET noticeably outperforms the baseline in Logical Reasoning (LR) question types for larger models Qwen3-32B (+0.0441) and LLaVa1.6-34B (+0.0735), demonstrating that MMCOMET makes retrieved commonsense knowledge more reliable, particularly for reasoning-heavy downstream tasks. Smaller models struggle more with MMCOMET alone, possibly due to their limited reasoning capabilities.

We note that performance varies across different model-category combinations, exhibiting differential sensitivities to the visually grounded commonsense knowledge from MMCOMET, highlighting the inherent heterogeneity of reasoning requirements in the dataset and the varying strengths of specific model types. Our adaptive routing mechanism is designed to exploit this heterogeneity by conditionally activating knowledge paths for each input based on predicted route utility and confidence. This dynamic routing not only selects the optimal path but also further mitigates extreme performance degradation from the rigid application of MMCOMET, resulting higher overall accuracies and mitigated effects from categories that do not benefit as much from commonsense knowledge, such as heavy perception-based tasks. This underscores the importance of dynamically bypassing knowledge paths when augmentation proves detrimental rather than beneficial to the downstream task.

\newpage
\clearpage
\section{Prompt Templates}
\label{appendix:prompts}
Tables \ref{tab:prompt_vcr_baseline}-\ref{tab:prompt_imagecaptioning_mmcomet} provide all the prompt templates for Visual Commonsense Reasoning, Visual Question Answering, Image Editing, and Image Captioning.


\captionof{table}{Baseline Prompt Template (VCR)}
\label{tab:prompt_vcr_baseline}
\begin{tcolorbox}[promptbox, breakable, title={Baseline Prompt Template (VCR)}]
You are given an image, a question, and four candidate rationales.\\

Choose the rationale that is best supported by the image and that most plausibly explains the situation in the question.\\
When deciding:\\
- Prioritise visible evidence in the image.\\
- Prefer the choice that best matches the people, objects, actions, and scene context.\\
- Reject choices that are too generic, unsupported, or contradicted by the image.\\
- Do not choose a rationale just because it sounds generally plausible.\\

\{fewshot\}\\

Question:\\
\{question\}\\
Candidate rationales:\\
(A) \{r0\}\\
(B) \{r1\}\\
(C) \{r2\}\\
(D) \{r3\}\\

Think silently about which option is most consistent with the image, then answer with the letter only.\\
Answer:
\end{tcolorbox}

\setlength{\abovecaptionskip}{0pt}
\setlength{\belowcaptionskip}{0pt}
\captionof{table}{TextKG Prompt Template (VCR)}
\label{tab:prompt_vcr_textkg}
\begin{tcolorbox}[promptbox, breakable, title={TextKG Prompt Template (VCR)}]
You are given an image, a question, four candidate rationales, and retrieved commonsense knowledge.\\

Your goal is to choose the rationale that is best supported by the image and most plausibly explains the situation. Use the retrieved knowledge only as weak background knowledge.
If the knowledge is irrelevant, vague, or not clearly supported by the image, ignore it.
Always prioritise the image and the candidate rationales over the retrieved knowledge.\\

Retrieved commonsense knowledge:\\
\{kg\_list\} \\
When deciding:\\
- First use the image to identify the most likely situation.\\
- Then compare all four rationales carefully.\\
- Prefer the rationale that is most specific to the visible scene.\\
- Do not choose a rationale only because it matches the retrieved knowledge.\\

\{fewshot\}\\

Question:\\
\{question\}\\
Candidate rationales:\\
(A) \{r0\} \\
(B) \{r1\} \\
(C) \{r2\} \\
(D) \{r3\} \\

Think silently about which option is most consistent with the image, then answer with the letter only.\\
Answer:
\end{tcolorbox}

\setlength{\abovecaptionskip}{0pt}
\setlength{\belowcaptionskip}{0pt}
\captionof{table}{MMCOMET Prompt Template (VCR)}
\label{tab:prompt_vcr_mmcomet}

\begin{tcolorbox}[promptbox, title={MMCOMET Prompt Template (VCR)}]
You are given an image, a question, four candidate rationales, and retrieved multimodal commonsense knowledge from visually similar situations.\\

Choose the rationale that is best supported by the image and most plausibly explains the situation.\\
The retrieved multimodal knowledge may help only when it matches the visible scene.\\
If it is generic, weakly related, or conflicts with the image, ignore it.
Always prioritise the image over the retrieved knowledge.\\

Retrieved multimodal commonsense knowledge:\\
\{kg\_list\} \\

When deciding:\\
- Use the image as the main evidence.\\
- Compare all four rationales, not just the first plausible one.\\
- Prefer the rationale that best explains the specific scene, actions, and relationships in the image.\\
- Avoid generic commonsense answers that are not clearly grounded in the image.\\

\{fewshot\}\\

Question:\\
\{question\}\\

Candidate rationales:\\
(A) \{r0\} \\
(B) \{r1\} \\
(C) \{r2\} \\
(D) \{r3\} \\

Think silently about which option is most consistent with the image, then answer with the letter only.\\
Answer:

\end{tcolorbox}


\captionof{table}{Baseline Prompt Template (VQA)}
\label{tab:prompt_vqa_baseline}
\begin{tcolorbox}[promptbox, title={Baseline Prompt Template (VQA)}]
You are given an image and a question.\\

Your task is to answer the question based only on the visual content.\\

\{fewshot\}\\

Question:\\
\{question\}\\

Answer with a short, concise response (1-3 words).\\
Answer:
\end{tcolorbox}

\captionof{table}{TextKG Prompt Template (VQA)}
\label{tab:prompt_vqa_textkg}
\begin{tcolorbox}[promptbox, title={TextKG Prompt Template (VQA)}]
You are given an image, a question, and retrieved commonsense knowledge.\\

Your task is to answer the question based primarily on the image.\\

Use the retrieved knowledge only as weak background information.
If it is irrelevant, vague, or not supported by the image, ignore it.\\

Retrieved commonsense knowledge:\\
\{kg\_list\}\\

When answering:\\
- First identify the relevant visual evidence in the image.\\
- Then consider whether the knowledge is helpful.\\
- Do not answer based only on the knowledge.\\
- Prefer answers that are directly grounded in the image.\\

\{fewshot\}\\

Question:\\
\{question\}\\

Answer with a short, concise response (1-3 words).\\
Answer:
\end{tcolorbox}

\captionof{table}{MMCOMET Prompt Template (VQA)}
\label{tab:prompt_vqa_mmcomet}
\begin{tcolorbox}[promptbox, title={MMCOMET Prompt Template (VQA)}]
You are given an image, a question, and retrieved multimodal commonsense knowledge from visually similar situations.\\

Your task is to answer the question based primarily on the image.\\

The retrieved multimodal knowledge may help only when it matches the visible scene.\\
If it is generic, weakly related, or conflicts with the image, ignore it.\\

Retrieved multimodal commonsense knowledge:\\
\{kg\_list\}\\

When answering:\\
- Use the image as the main source of evidence.\\
- Consider the knowledge only if it helps explain the specific situation in the image.\\
- Avoid generic answers that are not grounded in the image.\\
- Prefer answers that reflect the actual scene, objects, and interactions.\\

\{fewshot\}\\

Question:\\
\{question\}\\

Answer with a short, concise response (1-3 words).\\
Answer:
\end{tcolorbox}

\newpage
\captionof{table}{Baseline Prompt Template (Image Editing)}
\label{tab:prompt_imageediting_baseline}
\begin{tcolorbox}[promptbox, title={Baseline Prompt Template (Image Editing)}]
Background:\\
    You are an advanced image editing assistant designed to edit images based on given instructions. \\
    Your task is to apply the most appropriate visual modification to the input image based on the editing instruction provided.\\
    
    Input Information:\\
    Image: [input image]\\
    Editing Instruction: \{instruction\}
\end{tcolorbox}

\caption{MMCOMET Prompt Template (Image Editing)}
\label{tab:prompt_imageediting_mmcomet}
\begin{tcolorbox}[promptbox, title={MMCOMET Prompt Template (Image Editing)}]
Background:\\
    You are an advanced image editing assistant designed to edit images based on given instructions.\\
    Your task is to apply the most appropriate visual modification to the input image based on the editing instruction and the provided commonsense knowledge.\\
    Explicit commonsense knowledge consists of statements retrieved from a commonsense knowledge source, representing the beginning (head) and the outcome (tail) of a commonsense event.\\
    Each statement is retrieved based on its relevance to both the input image and the editing instruction.\\
    Your objective is to integrate the commonsense knowledge with the editing instruction to apply the most appropriate and contextually grounded visual modification.\\
    
    Input Information:\\
    Image: [input image]\\
    
    Explicit Commonsense Knowledge:\\
    Using the following knowledge graph in head-tail format: \{head\} $\rightarrow$ \{tail\}\\
    
    Editing Instruction: \{instruction\}
\end{tcolorbox}

\captionof{table}{Baseline Prompt Template (Image Captioning)}
\label{tab:prompt_imagecaptioning_baseline}
\begin{tcolorbox}[promptbox, title={Baseline Prompt Template (Image Captioning)}]
Write a natural one-sentence caption for the image.\\

\{fewshot\}\\

Caption:
\end{tcolorbox}

\caption{TextKG Prompt Template (Image Captioning)}
\label{tab:prompt_imagecaptioning_textkg}
\begin{tcolorbox}[promptbox, title={TextKG Prompt Template (Image Captioning)}]
You are given an image and some retrieved commonsense hints.\\
    Write a natural one-sentence caption for the image.\\
    Use the image as the primary source of information.\\
    Use the commonsense hints only if they are clearly supported by the image.\\
    Do not include details that are uncertain, speculative, or not visible.\\
    
    \{fewshot\}\\
    
    Retrieved commonsense hints:\\
    \{kg\_list\}\\
    
    Caption:
\end{tcolorbox}

\captionof{table}{MMCOMET Prompt Template (Image Captioning)}
\label{tab:prompt_imagecaptioning_mmcomet}
\begin{tcolorbox}[promptbox, title={MMCOMET Prompt Template (Image Captioning)}]
You are given an image and some retrieved multimodal commonsense hints.\\
    Write a natural one-sentence caption for the image.\\
    Use the image as the primary source of information.\\
    Use the commonsense hints only if they are clearly supported by the image.\\
    Do not include details that are uncertain, speculative, or not visible.\\
    
    \{fewshot\}\\
    
    Retrieved multimodal commonsense hints:\\
    \{kg\_list\}\\
    
    Caption:
\end{tcolorbox}

\clearpage

\end{document}